\documentclass{article}

\PassOptionsToPackage{numbers}{natbib}

\usepackage[preprint]{neurips_2026}

\usepackage[utf8]{inputenc} % allow utf-8 input
\usepackage[T1]{fontenc}    % use 8-bit T1 fonts
\usepackage{url}            % simple URL typesetting
\usepackage{booktabs}       % professional-quality tables
\usepackage{amsfonts}       % blackboard math symbols
\usepackage{nicefrac}       % compact symbols for 1/2, etc.
\usepackage{microtype}      % microtypography
\usepackage{xcolor}         % colors
\usepackage[pagebackref,breaklinks,colorlinks]{hyperref}

\usepackage{graphicx}
\usepackage{subcaption}
\usepackage{listings}
\usepackage{pgfplots}
\pgfplotsset{compat=1.18}
\usepackage{tikz}
\usetikzlibrary{fit}
\usepackage{marvosym}
\usepackage{pgfkeys}
\usepgfplotslibrary{statistics}
\usetikzlibrary{positioning, backgrounds, patterns, shapes.geometric, arrows.meta, calc}

\definecolor{visualblue}{RGB}{66, 133, 244}
\definecolor{textorange}{RGB}{234, 67, 53}
\definecolor{linkblue}{RGB}{0,0,180}
\definecolor{navyblue}{RGB}{0,38,84}
\definecolor{royalblue}{RGB}{28,69,135}

\usepackage{amsmath}
\usepackage{amssymb}
\usepackage{mathtools}
\usepackage{amsthm}
\usepackage{multirow}

\usepackage{tcolorbox}
\usepackage{fontawesome5}
\newtcolorbox{insight}[1][]{
  colback=royalblue!5!white,
  colframe=royalblue!95!black,
  fonttitle=\bfseries,
  title={\faLightbulb\ Key Insight},
  boxrule=1pt,
  arc=2mm,
  left=3pt,
  right=3pt,
  top=3pt,
  bottom=3pt,
  #1
}
\newtcolorbox{finding}[1][]{
  colback=green!5!white,
  colframe=green!50!black,
  fonttitle=\bfseries,
  title={\faCheckCircle\ Finding},
  boxrule=1pt,
  arc=2mm,
  left=3pt,
  right=3pt,
  top=3pt,
  bottom=3pt,
  #1
}
\newtcolorbox{observation}[1][]{
  colback=orange!5!white,
  colframe=orange!75!black,
  fonttitle=\bfseries,
  title={\faEye\ Observation},
  boxrule=1pt,
  arc=2mm,
  left=3pt,
  right=3pt,
  top=3pt,
  bottom=3pt,
  #1
}

\usepackage{enumitem}
\setlist[enumerate]{
  itemsep=1mm,
  topsep=1mm,
  parsep=0mm,
  partopsep=0mm,
  leftmargin=*
}
\setlist[itemize]{
  itemsep=1mm,
  topsep=1mm,
  parsep=0mm,
  partopsep=0mm,
  leftmargin=*
}

\usepackage{titlesec}
\titlespacing*{\paragraph}
{0pt}
{0mm}
{2mm}

\usepackage[capitalize,noabbrev]{cleveref}
\usepackage{xspace}

\theoremstyle{plain}

\theoremstyle{definition}

\theoremstyle{remark}

\newcommand\mypara[1]{\vspace{0.5mm}\noindent\textbf{#1}}

\usepackage[textsize=tiny]{todonotes}

\hypersetup{
  colorlinks   = true, %Colours links instead of ugly boxes
  urlcolor     = blue!30!black, %Colour for external hyperlinks
  linkcolor    = red, %Colour of internal links
  citecolor   = green!50!black %Colour of citations
}

\title{\emph{Not Another Text Benchmark:} \\ Putting the ``Visual'' Back in Visual Question Answering for Large Video Models}

\author{%
  \textbf{Rwiddhi Chakraborty}$^{1}$ \quad
  \textbf{Yinong Oliver Wang}$^{2}$ \quad
  \textbf{Cheng Zhang}$^{3}$ \\
  \textbf{Fan Bai}$^{4}$ \quad
  \textbf{Zhuoran Yu}$^{5}$ \quad 
  \textbf{Michael Kampffmeyer}$^{6}$ \\
  \textbf{Yong Jae Lee}$^{5}$ \quad
  \textbf{Fernando De la Torre}$^{2}$ \quad
  \textbf{Robert Jenssen}$^{6}$ \\ [1ex]
  $^{1}$ University of Copenhagen \quad
  $^{2}$ Carnegie Mellon University \quad
  $^{3}$ Texas A\&M University \\
  $^{4}$ Johns Hopkins University \quad
  $^{5}$ University of Wisconsin-Madison \\
  $^{6}$ UiT The Arctic University of Norway
}
\begin{document}

\maketitle

\begin{abstract}
Large video models have exhibited impressive performance on a wide range of visual question answering tasks, owing to the rise of powerful, pretrained text and vision encoders. The usefulness of such models have also been demonstrated on a wide range of benchmarks, with an important caveat - the dominant approach in these benchmarks evaluates multiple choice reasoning via \textit{text} options. This is a natural way to test text-based reasoning in these models, and has led to significant insights regarding model behavior in the community. In this work, we ask a different question - what happens when the evaluation modality is \textit{visual}, rather than text? We introduce three new vision-centric evaluation benchmarks in \textit{temporal frame retrieval}, \textit{video future prediction}, and \textit{causal memory distortion}, all designed around evaluating visual understanding capabilities in large video models. Our approach complements the existing approaches to evaluate video understanding in frontier models. We show that current frontier models exhibit significant weakness when attempting to reason through visual queries, rather than text. We conclude with an extended analysis section that provides pointers for future improvements in visual understanding for large video models\footnote{ Project page: \href{https://rwchakra.github.io/not-another-text-benchmark/}{https://rwchakra.github.io/not-another-text-benchmark/}}.

\end{abstract}

\section{Introduction}

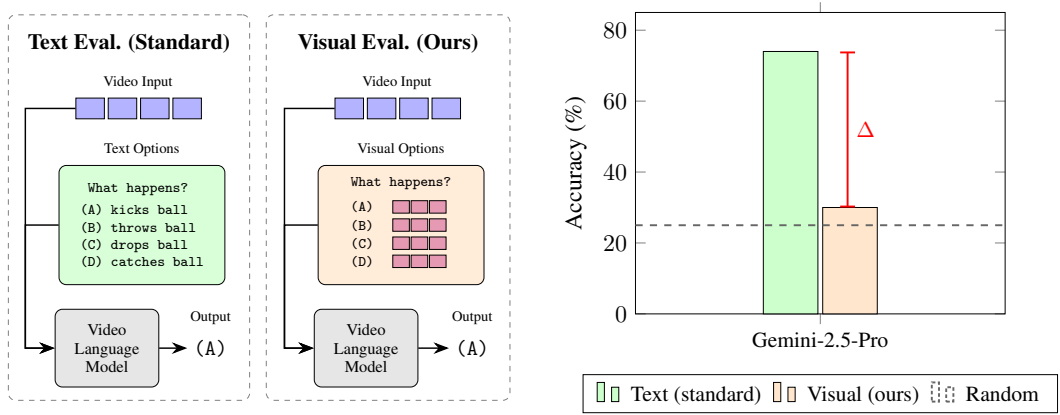
\begin{figure}[t!]
    \centering
    \begin{minipage}[c]{0.48\columnwidth}
        \centering
        \resizebox{\linewidth}{!}{\begin{tikzpicture}[
    node distance=1cm and 1.5cm,
    >={Stealth[length=3mm]},
    box/.style={draw, rounded corners, minimum width=2.2cm, minimum height=1cm, align=center, font=\small},
    modelbox/.style={box, fill=gray!20, minimum width=2cm, minimum height=1.5cm},
    framebox/.style={draw, minimum width=0.55cm, minimum height=0.4cm, fill=blue!30},
    smallframe/.style={draw, minimum width=0.32cm, minimum height=0.24cm, fill=purple!40},
    textmcq/.style={draw, rounded corners, fill=green!15, minimum width=3.2cm, minimum height=2.3cm, align=left, font=\footnotesize},
    vismcq/.style={draw, rounded corners, fill=orange!15, minimum width=3.2cm, minimum height=2.3cm, align=left, font=\footnotesize},
    label/.style={font=\bfseries\large},
]

% ============ LEFT SIDE: Standard Paradigm ============
\node[label] (left_title) at (-2.25, 3.5) {Text Eval. (Standard)};

% Video input (sequence of frames)
\node[framebox] (v1_l) at (-3, 2.25) {};
\node[framebox, right=0.06cm of v1_l] (v2_l) {};
\node[framebox, right=0.06cm of v2_l] (v3_l) {};
\node[framebox, right=0.06cm of v3_l] (v4_l) {};
\node[above=0.05cm of v2_l, font=\footnotesize, xshift=0.3cm] {Video Input};

% Text MCQ prompt
\node[textmcq] (prompt_l) at (-2, 0) {
    \texttt{ What happens?}\\[3pt]
    \texttt{(A) kicks ball}\\
    \texttt{(B) throws ball}\\
    \texttt{(C) drops ball}\\
    \texttt{(D) catches ball}
};
\node[above=0.05cm of prompt_l, font=\footnotesize] {Text Options};

% Model - moved right to add gap from bounding box
\node[modelbox] (model_l) at (-2.675, -2.375) {Video\\Language\\Model};

% Output
\node[font=\large\ttfamily] (output_l) at (-0.675, -2.375) {(A)};
\node[above=0.05cm of output_l, font=\footnotesize] {Output};

% Arrows
% Video input arrow - exits west, goes down to joining point
\draw[->, thick] (v1_l.west) -- (-4.25, 2.25) |- (model_l.west);
% Text prompt arrow - exits west, joins and goes to model west
\draw[->, thick] (prompt_l.west) -- (-4.25, 0) |- (model_l.west);
% Model to output
\draw[->, thick] (model_l.east) -- (output_l.west);

% Bounding box for left
\begin{scope}[on background layer]
    \node[draw=gray, dashed, rounded corners, fit=(left_title)(v1_l)(v4_l)(prompt_l)(output_l)(model_l), inner sep=0.25cm] (leftbox) {};
\end{scope}

% ============ RIGHT SIDE: Our Paradigm ============

\node[label] (right_title) at (2.75, 3.5) {\;\;Visual Eval. (Ours)\;\;};

% Video input (sequence of frames) - same relative position as left
\node[framebox] (v1_r) at (2, 2.25) {};
\node[framebox, right=0.06cm of v1_r] (v2_r) {};
\node[framebox, right=0.06cm of v2_r] (v3_r) {};
\node[framebox, right=0.06cm of v3_r] (v4_r) {};
\node[above=0.05cm of v2_r, font=\footnotesize, xshift=0.3cm] {Video Input};

% Visual MCQ box - same relative position as left prompt
\node[vismcq] (prompt_r) at (3, 0) {};
\node[above=0.05cm of prompt_r, font=\footnotesize] {Visual Options};

% Question text inside visual MCQ - positioned relative to prompt_r center
\node[font=\footnotesize] at ([yshift=0.8cm]prompt_r.center) {\texttt{What happens?}};

% Option A - sequence of small frames
\node[font=\footnotesize] (optA) at ([xshift=-0.75cm, yshift=0.35cm]prompt_r.center) {\texttt{(A)}};
\node[smallframe] (a1) at ([xshift=0cm, yshift=0.35cm]prompt_r.center) {};
\node[smallframe, right=0.02cm of a1] (a2) {};
\node[smallframe, right=0.02cm of a2] (a3) {};

% Option B
\node[font=\footnotesize] (optB) at ([xshift=-0.75cm, yshift=0.0cm]prompt_r.center) {\texttt{(B)}};
\node[smallframe] (b1) at ([xshift=0cm, yshift=0.0cm]prompt_r.center) {};
\node[smallframe, right=0.02cm of b1] (b2) {};
\node[smallframe, right=0.02cm of b2] (b3) {};

% Option C
\node[font=\footnotesize] (optC) at ([xshift=-0.75cm, yshift=-0.35cm]prompt_r.center) {\texttt{(C)}};
\node[smallframe] (c1) at ([xshift=0cm, yshift=-0.35cm]prompt_r.center) {};
\node[smallframe, right=0.02cm of c1] (c2) {};
\node[smallframe, right=0.02cm of c2] (c3) {};

% Option D
\node[font=\footnotesize] (optD) at ([xshift=-0.75cm, yshift=-0.7cm]prompt_r.center) {\texttt{(D)}};
\node[smallframe] (d1) at ([xshift=0cm, yshift=-0.7cm]prompt_r.center) {};
\node[smallframe, right=0.02cm of d1] (d2) {};
\node[smallframe, right=0.02cm of d2] (d3) {};

% Model - same relative position as left
\node[modelbox] (model_r) at (2.325, -2.375) {Video\\Language\\Model};

% Output
\node[font=\large\ttfamily] (output_r) at (4.375, -2.375) {(A)};
\node[above=0.05cm of output_r, font=\footnotesize] {Output};

% Arrows
% Video input arrow - exits west, goes down to joining point
\draw[->, thick] (v1_r.west) -- (0.75, 2.25) |- (model_r.west);
% Visual prompt arrow - exits west, joins and goes to model west
\draw[->, thick] (prompt_r.west) -- (0.75, 0) |- (model_r.west);
% Model to output
\draw[->, thick] (model_r.east) -- (output_r.west);

% Bounding box for right - include ALL elements like left side
\begin{scope}[on background layer]
    \node[draw=gray, dashed, rounded corners, fit=(right_title)(v1_r)(v4_r)(prompt_r)(output_r)(model_r), inner sep=0.25cm] (rightbox) {};
\end{scope}

\end{tikzpicture}}
    \end{minipage}
    \hfill
    \begin{minipage}[c]{0.48\columnwidth}
        \centering
        \resizebox{\linewidth}{!}{\begin{tikzpicture}[
    node distance=1cm and 1.5cm,
    >={Stealth[length=3mm]},
    box/.style={draw, rounded corners, minimum width=2.2cm, minimum height=1cm, align=center, font=\small},
    modelbox/.style={box, fill=orange!20, minimum width=2.8cm, minimum height=1.2cm},
    framebox/.style={draw, minimum width=0.55cm, minimum height=0.4cm, fill=blue!30},
    smallframe/.style={draw, minimum width=0.32cm, minimum height=0.24cm, fill=purple!40},
    label/.style={font=\bfseries\large},
]
\begin{axis}[
    ybar,
    bar width=0.8cm,
    width=7cm,
    height=6cm,
    ylabel={Accuracy (\%)},
    ylabel style={font=\normalsize},
    symbolic x coords={Gemini-2.5-Pro},
    xtick=data,
    xticklabel style={font=\small},
    ymin=0,
    ymax=85,
    ytick={0,20,40,60,80},
    legend style={
        at={(0.5,-0.20)},
        anchor=north,
        legend columns=3,
        column sep=1mm,
        inner ysep=1mm,
        inner xsep=2mm,
        font=\small,
        nodes={text height=1.25mm, text depth=0mm}
    },
    ymajorgrids=false,
    enlarge x limits=0.5,
    clip=false,
    clip mode=individual,
]
% Text Accuracy (standard) -- drawn on the left
\addplot[fill=green!20, draw=black] coordinates {
    (Gemini-2.5-Pro,74)
};
% Visual Accuracy (ours) -- drawn on the right
\addplot[fill=orange!20, draw=black] coordinates {
    (Gemini-2.5-Pro,30.0)
};
% Delta interval line (vertical, between the two bar heights)
\draw[|-|, thick, red]
    ([xshift=0.4cm]axis cs:Gemini-2.5-Pro,30.0) --
    ([xshift=0.4cm]axis cs:Gemini-2.5-Pro,74.0)
    node[midway, right, red, font=\small] {$\Delta$};
% 25% Random Baseline (horizontal dashed line)
\draw[thick, dashed, color=black!60]
    ({rel axis cs:0,0}|-{axis cs:Gemini-2.5-Pro,25}) --
    ({rel axis cs:1,0}|-{axis cs:Gemini-2.5-Pro,25});
\addlegendentry{Text (standard)}
\addlegendentry{Visual (ours)}
\addlegendimage{thick, dash pattern=on 1.5pt off 1pt, color=black!60}
\addlegendentry{Random}
\end{axis}
\end{tikzpicture}}
    \end{minipage}
    \caption{\small A simple illustration of the usefulness of our visual evaluation suite. Traditional evaluation suites in video understanding tasks use text as the evaluation modality (left). We introduce a dataset that evaluates visual understanding in large video models through the visual modality (right). On the temporal frame retrieval task in our dataset, Gemini 2.5 Pro, a frontier video model, performs significantly worse when the answer choices are visual than when they are text (the \textcolor{red}{$\Delta$} sign signifies the drop in accuracy, over 40 percentage points). This demonstrates a gap in the visual reasoning capabilities of large video models.}
    \label{fig:teaser}
    \vspace{-5mm}
\end{figure}

The rise of large video models in recent years is one of the most interesting developments in multimodal artifical intelligence \cite{wiedemer2025video, deitke2025molmo, wang2024internvideo2,yenchenlin2024video}. These models typically follow the architectural primitives of large vision-language models \cite{bai2025qwen3, liu2023visual, li2023blip, ataallah2024minigpt4}, where pretrained vision and text encoders are aligned such that tokens in both modalities may be interleaved to be interpreted as inputs to the generative decoder. This approach, coupled with training on large amounts of data, has led to impressive visual understanding capabilities of these models~\cite{tang2025video}. 
% However, there are several works which have shown that the apparent capabilities of these models are better attributed to a variety of \yj{language-based} shortcuts, rather than true visual reasoning [cite]. \yj{It would be good to give an example of such a shortcut.} These findings are consistent across both vision-language, and video-language models [cite]. 

We notice an exclusive design choice in the benchmarks that evaluate video understanding - examples include TemporalBench ~\cite{Cai2024TemporalBenchBF}, TempCompass ~\cite{liu2024tempcompass}, Vinoground ~\cite{zhang2024vinoground}, MVBench ~\cite{li2024mvbench}, and LVBench~\cite{wang2025lvbench}. They construct multiple choice questions \textit{exclusively in text}. While using text as an evaluation modality is valid for the visual question answering (VQA) task, we are motivated to create a more direct evaluation suite for visual understanding, i.e., creating datasets where the evaluation modality is visual. This approach has two benefits: first, it gives us a complementary view in evaluating the same VQA datasets on the same models in a different modality than text. In this way, the visual modality appears as a useful control. Second, this approach allows us to investigate whether frontier video models could demonstrate inflated VQA performance by using language cues in the answer choices. There exists evidence that they do so ~\cite{agrawal2018don, ko2023large}. In fact, in TemporalBench, the authors in an ablation find that large language models (LLMs) will often use a ``centralized'' description of the options as a cue to make the final decision. Some of these results thus indicate a greater need for direct visual evaluation of such models. While we do not directly tackle language shortcuts in this work, we are able to shed light on certain aspects of language-based biases in large video models through our novel evaluation design. We illustrate our motivation through a simple experiment in \Cref{fig:teaser}. On a task where we prompt a state-of-the-art model (Gemini-2.5-Pro~\cite{comanici2025gemini}) to answer what happens after (or before) a particular action in the input video, the model performs well when the multiple choice options are in text. However, when we replace these text options with \textit{visual options} (from the actual input video), the performance drop is stark. This phenomenon of incorrect visual grounding will be a recurrent theme in this paper. Targeting large video-language models, we create our own novel evaluation suite based on this theme.
% While these benchmarks have effectively identified critical flaws and useful insights, if we wish to evaluate true visual capabilities in models, and given strong evidence of textual shortcuts in these models [cite], it is natural that a fair evaluation for visual capabilities should be performed on visual MCQ tasks. 
% While these benchmarks have effectively identified critical flaws and offered useful insights, a fair assessment of visual capability-especially in light of strong evidence of textual shortcuts [cite]-should be conducted on visual multiple-choice question (MCQ) tasks with video as options. 

While recent works in cross-modal consistency have hinted at this problem ~\cite{van2025same, zhang2024cross, nguyen2025see}, there are key differences: first, the current approach to check cross-modal consistency is to take images of data in other modalities, e.g., taking a picture of a graph, instead of taking the actual graph. In our work, vision is not a proxy for any other modality. It is the modality of evaluation. Second, none of the existing approaches address inconsistencies in video understanding. 
% This is where our contributions lie.

% Naturally, the challenges of operating in a different domain (in our case, vision) lead to quite different problems that we look to solve. 
In each of our tasks, the input is a video (a sequence of frames), and a set of \textit{visual} questions. These questions are either in the format of multiple choices (where each choice is a sequence of frames taken from the context video), or a single, binary decision-making problem, with a single frame as input. We choose three tasks:
\begin{itemize}
    \item \textit{Temporal frame retrieval}: the model is prompted to choose the correct sequence of frames after or before an anchor event. This task assesses frame retrieval and temporal reasoning of events.
    \item \textit{Video future prediction}: given query frames, the model is prompted to choose the correct subsequent future frames. This task tests understanding of physical and causal dynamics across time.
    \item {\textit{Causal memory distortion}}: inspired by an influential result in cognitive science~\cite{Strickland2011EventCE}, we evaluate memory distortion in large video-language models, which tests hallucination of a contact moment (e.g. throwing, kicking something) when the corresponding frame is removed from the input video. Our experimental design closely follows this original study, which shows that people often hallucinate the moment of contact when subsequent footage implies it, but not otherwise. To the best of our knowledge, this is the first evaluation of causal memory distortion in large video-language models.
    % - Inspired from an influential result in cognitive science \cite{strickland2011event}, we evaluate large video models in memory distortion, i.e. given an input video showing contact (e.g. throwing something, kicking something), does the model hallucinate the contact frame if the contact frame is removed in the video? This is an interesting question, since in the original work on humans, the authors demonstrated that humans falsely remembered the moment of contact if subsequent footage implied it, as opposed to not falsely remembering the moment if subsequent footage did not imply it. To the best of our knowledge, ours is the first work evaluating causal memory distortion in large multimodal models.
\end{itemize}

% \begin{figure}[t]
%     \centering
%     \resizebox{\linewidth}{!}{\input{figures/intro.tex}}
%     \caption{Traditional evaluation suites in video understanding tasks use text as the evaluation modality (left). We introduce a dataset that evaluates visual understanding in large video modalities through the visual modality(right).}
%     \label{fig:intro}
% \end{figure}

Our results reveal substantial weaknesses in these core visual tasks across state-of-the-art video-language models, introducing new aspects against recent claims that these models are zero-shot reasoners~\cite{wiedemer2025video}. We conclude with an extended analysis section, shedding further light on various aspects of visual understanding in frontier video models. In summary, our contributions include:

\begin{itemize}
    \item A novel benchmark of visual questions, beyond the current dominant approach of text-based questions in querying video understanding in large video models.
    \item Evaluating \textit{causal memory distortion} in large video models, a purely visual task, inspired from cognitive science. To the best of our knowledge, this study does not exist in the current large vision model (LVM) literature.
    \item Surprising results of poor model performance and extended analysis that provides insights on model failure modes, paving the way for future improvements in model performance.
\end{itemize}

\section{Related Work}

% In this section, we review existing literature along two fronts: \textit{shortcut learning in VLMs} and \textit{relevant benchmarks}.

\mypara{Visual Understanding in Video Models.}
Before the advent of large multi-modal models (LMMs) in video,
pretrained models such as CLIP ~\cite{radford2021learning} trained on video datasets exhibited high levels of single frame reliance \cite{lei2023revealing,buch2022revisiting}. In \cite{buch2022revisiting},
the authors present a CLIP-first approach to extracting a single, representative frame from the input video for a downstream video QA/retrieval task. This approach was found to be sufficient for downstream video QA and retrieval tasks. Similarly, shuffling frames in videos to create temporal classes reveal that most major benchmarks overwhelmingly contain static rather than temporal classes: solving these do not require temporal reasoning \cite{sevilla2021only}. Through our dataset design, we explicitly enforce the \textit{temporal} aspect of video understanding since all three tasks require the model to reason \textit{across} frames, rather than a single frame alone: whether it be for retrieval, memory distortion, or future prediction. With the introduction of video LMMs, in addition to the single frame bias, a large body of work showed that such models frequently rely on \textit{textual} shortcuts when making decisions \cite{thrush2022winoground,Ye_2021_AAAI,wang2023dataset,10.1609/aaai.v37i11.26590,levy2023guiding,kil2021discovering}. More recent multimodal LLMs (MLLMs) evaluations have focused on hallucination issues~\cite{tong2024eyes}. In this case, it seems even more relevant to design visual evaluation suites, to better ascertain the impact of language-based biases in model decision-making. 
While our work does not directly tackle shortcut learning, we demonstrate the benefits of using a visual evaluation suite to shed light on this problem.

\mypara{Relevant Benchmarks.}
 Several benchmarks have been proposed to evaluate the capabilities
of video LMMs \cite{yi2021benchmarking, Cai2024TemporalBenchBF,zhang2024vinoground,li2024mvbench,Cores2024LostIT, girdharcater}. However, the key approach underlying all these works is the choice of \textit{text} as the primary query modality. Although evaluating text-based reasoning is quite natural for certain VQA tasks, they do not provide a mechanism to directly evaluate visual reasoning and groundedness, i.e., reasoning through vision rather than text. Thus, our evaluation suite provides a more direct way to compare text-based and vision-based reasoning, and positions itself complementarily to existing approaches. The approach most similar to ours is in the evaluation of audio models \cite{nguyen2025see}, where the authors propose similar questions to evaluate audio models. However, there are key differences in our work: first, we operate in a different domain of video, and second, we note that the issue of memory distortion that we tackle here has not been explored in other works either. We believe that this task is a novel way to detect hallucinations in large video models. 

\mypara{Cross-modal consistency.}
Cross-modal consistency refers to evaluating the consistency of model behaviour under information streams from different modalities. The advent of MLLMs has made this issue particularly relevant, since truly multimodal learning requires models to synthesize information across complementary modalities. This is not a new phenomenon. There are works that document the difficulty of training and evaluating multimodal architectures ~\cite{wang2020makes}, while others train unimodal encoders in an alternative fashion to reduce multimodal interference ~\cite{zhang2024multimodal}. More recent works have shifted the focus from representation learning to inference, owing to the heavy cost of training MLMMs. These include evaluating consistency using input and text and images of the same text ~\cite{van2025same}, and in visual algorithmic reasoning ~\cite{park2025generalizing}. Using screenshots of different modalities to create proxy visual information is also common ~\cite{zhang2024cross, van2025same}. Our work exists in a different domain, i.e., video understanding, and does not use image as a proxy modality: we use visual information directly in our evaluation suite.

% In the next section, we outline the dataset construction and furnish details of the three tasks: \textit{(1) temporal frame retrieval, (2) video future prediction, and (3) causal memory distortion}.

\vspace{-2mm}
\section{The Dataset}

Video understanding should require temporal and causal abilities, while remaining grounded in the visual information supplied. These three competencies, although not comprehensive, are a necessary set of requirements that repeatedly arise in real video use cases, such as those in retrieval, monitoring, and summarization tasks. We accordingly design three visual-centric tasks that each targets one essential axis of video understanding while keeping the interface simple and comparable across models.

In this section, we present details of the dataset design for three visual tasks
tackled in this paper: \textit{(1) temporal frame retrieval, (2) video future prediction, and (3) causal memory distortion}.

\paragraph{The unifying view}

Each task presented in our benchmark is a unique instantiation of the overarching schema: each instance as an input video/video prefix, a visual query, and optionally visual answer candidates. In every case, the model must pick the correct answer based on \textit{visual} evidence. Based on the exact nature of visual reasoning, the particular task differs. We present the three tasks below.

\vspace{-2mm}
\subsection{Temporal Frame Retrieval}
\label{sec:dataset_tfr}

This task evaluates a model's ability to understand temporal context in video.
Given a video clip and an \textit{anchor action} as input, the model is required to select
the correct frames that occur either before or after the anchor action from a set of candidate frames. 
The task is purely visual: the model is required to select among visual options, not text.

We construct this dataset using video clips from the Charades dataset~\cite{sigurdsson2016hollywood}, which provides
temporally annotated human action sequences, e.g., \verb|opening a door|, \verb|tidying some clothes|.
Given a video clip, we select an \textit{anchor action} from its annotations and pose two
temporal reasoning questions:
(i) what happens \textbf{before} the anchor action, and
(ii) what happens \textbf{after} the anchor action.
Crucially, the model must retrieve the correct answer from a set of candidate \emph{visual} options
(i.e., a visual MCQ), rather than text descriptions. 

\begin{figure}[t!]
    \centering
    \begin{minipage}[t]{0.49\textwidth}
        \vspace{0pt}
        \centering
        \includegraphics[width=\linewidth]{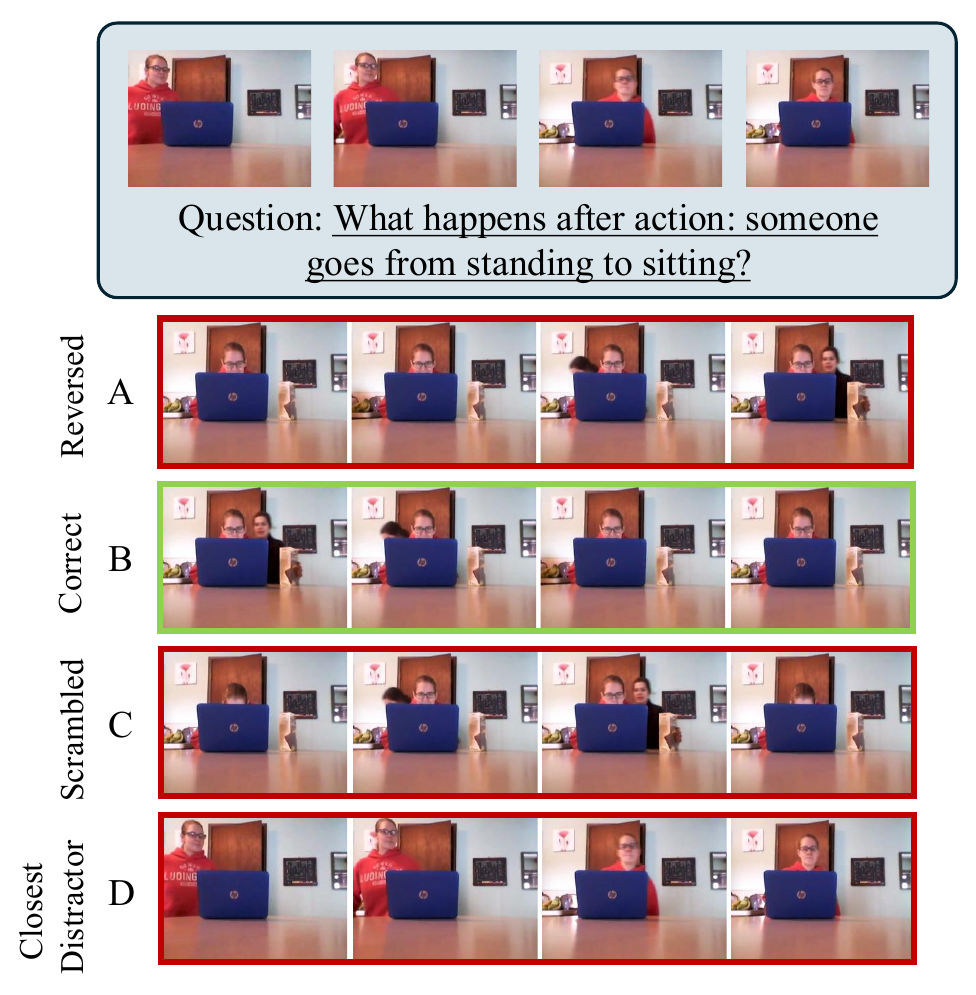}
        \caption{\small Temporal Frame Retrieval. Given a video and an action (e.g. ``someone goes from standing to sitting''), the model must select the correct sequence of frames that occur \textit{after} or \textit{before} the anchor action from four visual options. \textcolor[HTML]{9DC183}{Green} is correct, red incorrect.}
        \label{fig:temporal_retrieval}
    \end{minipage}
    \hfill
    \begin{minipage}[t]{0.49\textwidth}
        \vspace{1mm}
        \centering
        \includegraphics[width=\linewidth]{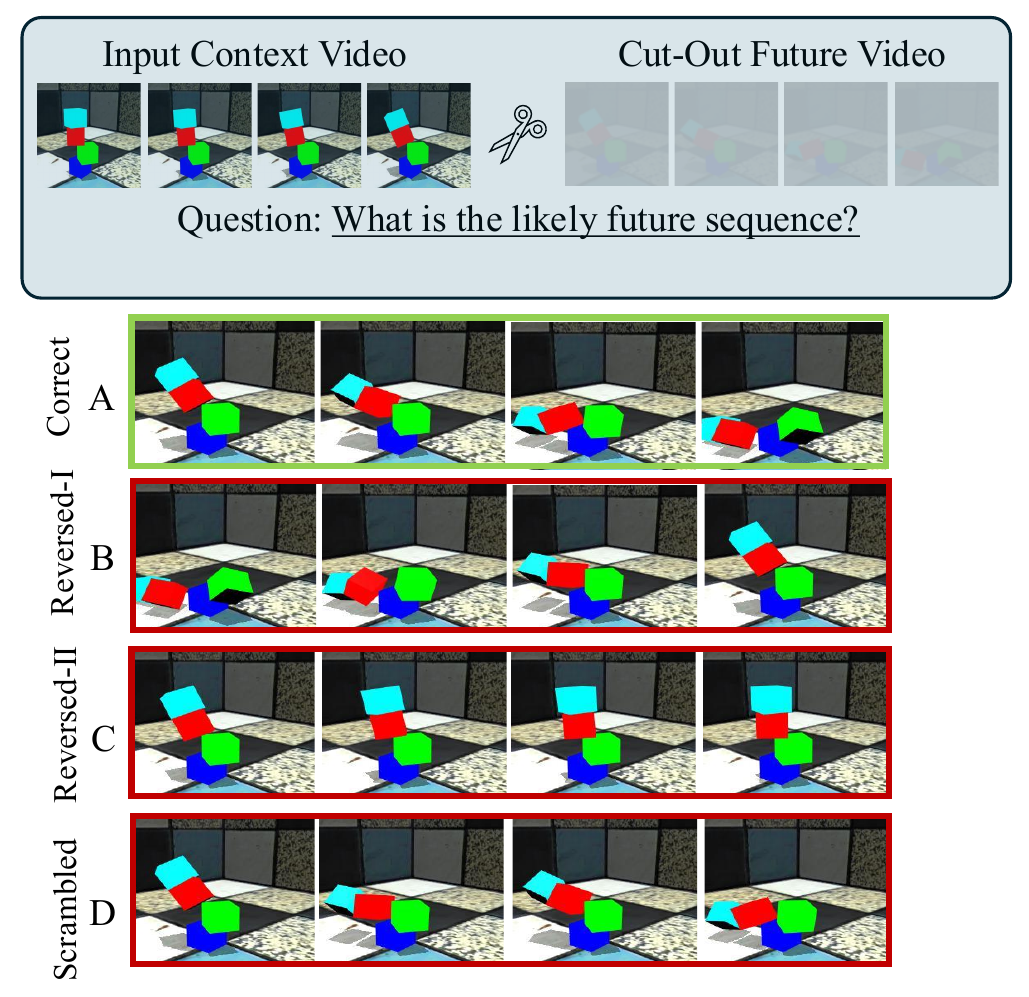}
        \vspace{-1mm}
        \caption{\small Video Future Prediction. Given a video input showing the initial context frames, the model must select the correct sequence of future frames from four visual options. \textcolor[HTML]{9DC183}{Green} is correct, red incorrect.}
        \label{fig:future_prediction}
    \end{minipage}
    % \vspace{-4mm}
\end{figure}

In our visual MCQ setup, each question is accompanied with four candidate options (as demonstrated in \cref{fig:temporal_retrieval}):
\begin{enumerate}
    \item \textbf{Correct}: the true temporal neighbor frames,
    \item \textbf{Scrambled}: frames sampled across the video and randomly shuffled,
    \item \textbf{Reversed}: the same span of frames as the correct option, played backward,
\item \textbf{Closest Distractor}: the false temporal neighbor frames, i.e. previous action to the anchor if the question is of type “after", and vice versa.
\end{enumerate}

There is only one correct answer per question, although future work could explore multiple correct options.
We employ a small set of preprocessing and filtering techniques to make sure we only include high-quality samples.\footnote{See \cref{app:dataset} in the appendix for data preprocessing details.}
We repurposed the Charades test set of 1862 videos and filtered down to 321 videos. 
After balancing the \textit{before} and \textit{after} questions, we constructed a total of 1108 questions. The correct answer 
for each question was assigned randomly among the four options. The dataset contains around 35,000 frames that we 
extract using this process. 

\subsection{Video Future Prediction}
\label{sec:dataset_vfp}

This task evaluates a model's ability to predict future events in a video from past context.
Given a video clip in which only an initial frame segment is visible (i.e. only a truncated video as input), the model must select the correct future frames from a set of candidate segments. This is different from the retrieval task, since the model does not have access to the entire video clip to make the decision. Instead, it has to reason based on the truncated context it receives as input. We construct this dataset using
video clips from the Shapestacks dataset~\cite{groth2018shapestacks}.
Shapestacks is a synthetic dataset created from a 3D simulator of blocks in a physical environment. The simulator creates instances of block configurations that are either stable or unstable, thus making it a natural testbed for the future prediction task.

First, we generate videos through the simulator, and for each selected video, extract the first eight frames as the \textit{context} frames
and the last eight frames as the \textit{future} frames.
Then, we generate multiple-choice questions for input videos, and
each question consists of four candidate options:
\begin{enumerate}
    \item \textbf{Correct}: the ground-truth future segment,
    \item \textbf{Reversed--I}: the ground-truth future frames presented in reverse order,
    \item \textbf{Reversed--II}: the input (context) frames presented in reverse order,
    \item \textbf{Scrambled}: the first and last frames are preserved, while the intermediate frames are randomly permuted.
\end{enumerate}

As in the previous dataset, each question contains exactly one correct answer, which is
randomly positioned among the four options.
In total, we finalize 300 video clips, resulting in 300 questions.
This process yields 12{,}000 extracted frames overall. We illustrate the task in \cref{fig:future_prediction}.

\begin{figure}[t]
    \centering
    \includegraphics[width=0.9\linewidth]{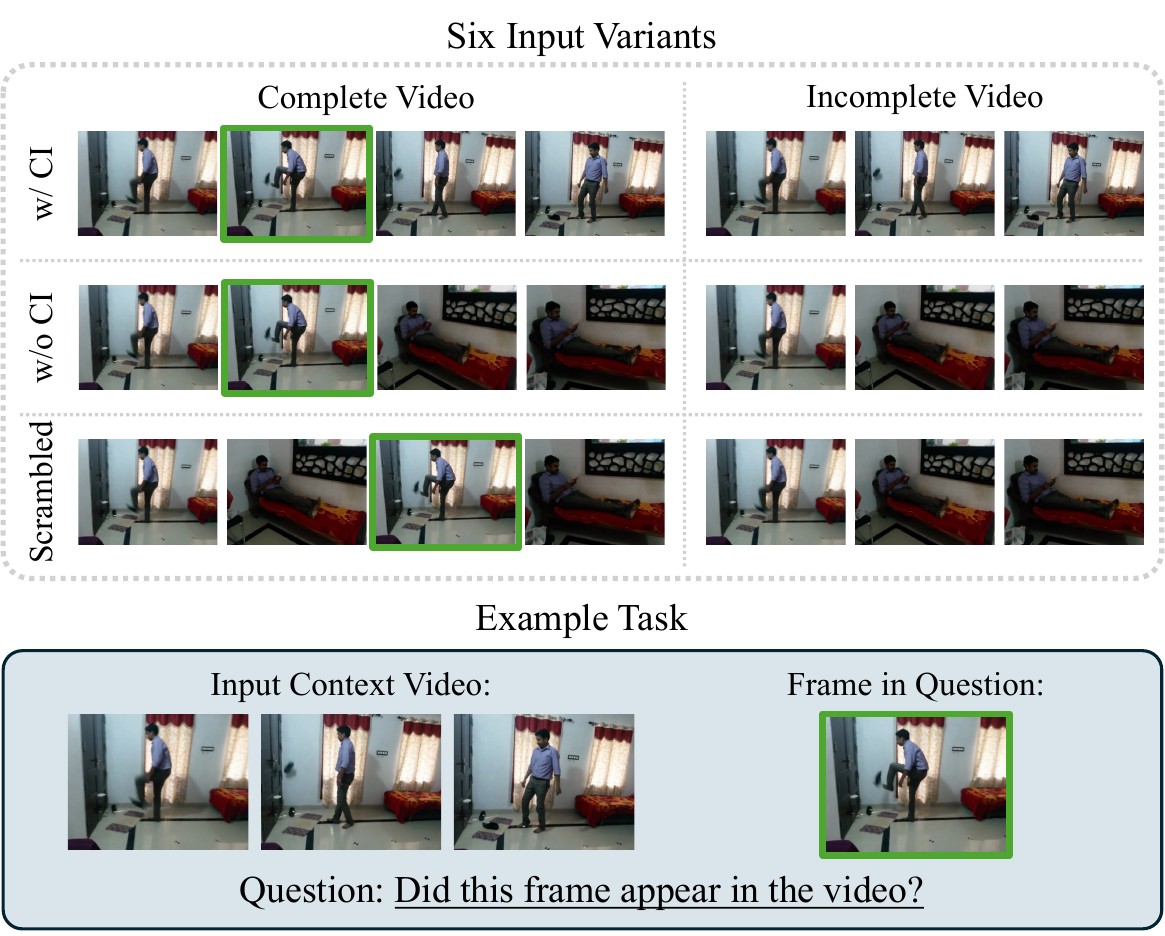}
    \caption{\small An illustration of the Causal Memory Distortion task. The contact frame (the moment the shoe leaves the foot) is highlighted in \textcolor[HTML]{9DC183}{green}. We denote w/CI--with causal implication, and w/o CI--without causal implication. The contact frame is present in the complete versions across the three conditions, but is absent in the incomplete versions.}
    \label{fig:causal_distortion}
    \vspace{-3mm}
\end{figure}

\subsection{Causal Memory Distortion}
\label{sec:dataset_cmd}

This task is inspired by an influential cognitive science experiment by \cite{Strickland2011EventCE}, which studies how humans form false memories about causal events in videos. 
% In light of recent developments of foundation models, we consider this finding as evidence of ``hallucination'' in human beings. 
In their setup, a video sequence contains a clear \textit{contact moment} (e.g., \verb|kicking something| or \verb|throwing something|). The video is then slightly altered to remove the contact moment, but subjects are shown a candidate ``contact moment'' frame from the original video. The study found that humans often falsely report having seen the contact moment in the altered video, effectively ``hallucinating'', when the remaining footage \textit{causally implies} the missing contact. Interestingly, this hallucination tendency in humans is less likely if the remaining footage does not causally follow the contact moment (e.g., footage of a ship sailing following the footage of a person about to kick a ball). 
To the best of our knowledge, we are the first to adapt this idea to curate a dataset that evaluates models' ability to detect causal inconsistencies in video sequences, i.e. evaluating whether models can indeed stay robust when contact frames are removed \textit{with causal implication}, as opposed to when they are not. 
We use the term ``causal memory distortion'' as the phenomenon when models exhibit similar hallucination about the removed ``contact'' frame.

To construct this dataset, we again use the Charades dataset. Since this task requires fine-grained control on the video editing, 
we manually selected videos from the Charades test set that satisfied the following three criteria: (i) the contact moment must be clear and long enough to be visible, (ii) the video must have a clear causal implication after the contact moment, and (iii) the video must also have a moment of clear non-causal implication. We do not require that the non-causal implication frames necessarily follow the contact moment immediately. 
After an extensive manual search, we finalized 20 videos that satisfied these criteria. While the dataset size is seemingly small, it is three times larger than the original work by \cite{Strickland2011EventCE}, which had only six videos. Further, the code to annotate more videos is extensible, and it is possible to add more videos beyond Charades for future work. 
% We reiterate the importance of the manual curation process here. 

With appropriate raw videos selected, we manually constructed the task setup, including the contact moment and relevant frames. We categorize two types of videos: complete, where the contact moment is not removed, and incomplete, where the contact moment is removed. Then, for each video type, three conditions were created (as illustrated in \cref{fig:causal_distortion}):
\begin{enumerate}
\item \textbf{With Causal Implication}: The video segment following the contact moment was preserved. 
\item \textbf{Without Causal Implication}: The video segment following the contact moment was replaced with a random segment causally unrelated to the contact moment.
\item \textbf{Scrambled}: The video segments were randomly shuffled, providing the necessary control. 
\end{enumerate}

In the end, we had six variants per original video (two video types, three conditions). During evaluation, the complete and incomplete video types 
were randomly assigned to each model. The task thus evaluates whether the model is more likely to hallucinate the contact moment when it is removed from the original video
but followed up with a clip that implies a causal connection. 

Our benchmark thus combines three visual evaluation datasets for evaluating frontier models, leading us to name it \textbf{Not Another Text Benchmark for Video (NATB-V)}\footnote{We agree that naming creativity is not the core contribution of this paper}.

\paragraph{A note on benchmark size and scaling}
The dataset currently has 641 videos. Temporal retrieval scales to any dataset with timestamped
descriptions; video future prediction scales via the physics simulator. The causal memory distortion task in particular is a bit difficult to scale, given the heavy manual curation required. However, our dataset has three times the number of videos from the original study. As we are not targeting a new state-of-the-art, we believe our sample size is reasonable for analyzing frontier models. Our 95\%
confidence intervals (bootstrap B=10000 and Wilson score) are tight, confirming sufficient dataset size
for reliable conclusions.

% The code to create these video variants can be used on other datasets as well, and we provide it, along with other details on the dataset creation process, in the supplementary material. An illustration of the task is provided in figure \ref{fig:causal_distortion}.

\section{Experiments}
% In this section, we outline the experimental setup, the models used, and the main results of this work. 

\subsection{Setup}
% \Cheng{can we introduce datasets, models, metrics separately?}
% \paragraph{Data} Since the question formats of are slightly different for each task, we evaluate them separately and present the results together. The appendix contains more information on the exact evaluation formats, prompt structures, and data formats.
\paragraph{Models} For our work, we use Gemini-2.5-Pro and GLM-5V-Turbo ~\cite{comanici2025gemini, hong2026glm} as the proprietary frontier video models, with LLaVA-OV-72B~\cite{lillava}, Qwen2.5-VL-72B, Qwen2.5-VL-32B~\cite{bai2025qwen25vltechnicalreport}, Qwen3-VL-32B~\cite{bai2025qwen3}, mPLUG-Owl3~\cite{yemplug}, InternVL3-78B~\cite{zhu2025internvl3}, Kimi-K2.6~\cite{team2026kimi}\footnote{No dedicated technical report is available for Kimi-K2.6, so we cite the report for Kimi-K2.5, whose multimodal architecture it extends. Official release post: \url{https://www.kimi.ai/ai-models/kimi-k2-6}, accessed October 26, 2026.}, and GLM-4.6V \cite{zeng2025glm} as the open-source frontier video models. We use fixed prompts for each task across the models to ensure consistency. We use models off the shelf through open-source downloads on Huggingface \cite{wolf2019huggingface}, or frontier model APIs.
\paragraph{Metrics} Unless otherwise noted, we present the accuracy of the models on all our tasks. For MCQ, the accuracy is measured from the selection of the ground truth option. For the Causal Memory Distortion task, the accuracy is measured based on whether the model predicts yes or no regarding the frame in question. We observe that the evaluated models follow instructions perfectly and always output the selected option or yes/no depending on the task.
\paragraph{Human baseline}
% To ensure that the tasks are not ill-posed and are indeed solvable, we also test ten human subjects on a subset of the dataset and consequently compare with the models. However, we would like to clarify here that performing a human evaluation and directly comparing with models in the current setup runs into a variety of scientific control issues - first, humans may look at the same video multiple times, while models get one forward inference pass. Second, humans have short-term memory, which gives them some semantic access to the data they see. The models do not have this form of memory - every forward pass is a fresh start. While some benchmarks do indeed present human studies, these are contentious, as the correct standards or designing scientifically sound human-model evaluations is an open question, and is an interesting direction of future work. Therefore,
% the human results here are only presented to demonstrate that the tasks are indeed solvable, and we make no claims on human-model performance comparisons. 
To verify that the tasks are well-posed and solvable, we tested ten human subjects on a subset of the dataset. However, directly comparing humans and models here raises scientific control issues: humans have short-term memory providing semantic access to prior data, whereas each model inference is a fresh start. Some benchmarks include human studies, but these remain contentious, as designing sound human-model evaluations is an open question and a direction for future work. We therefore present the human results only to demonstrate that the tasks are solvable, and make no claims about human-model performance comparisons.

We leave more information on the exact evaluation formats, prompt structures, and data formats in \cref{app:eval}.
\subsection{Main Results}

The main results are presented in~\Cref{tab:main_results}. Overall, the results demonstrate a surprisingly poor performance when probing on direct visual information. On the temporal frame retrieval task, we observe that all models perform poorly (more than half below random chance). On the future prediction task, we observe that Gemini significantly outperforms the remaining models, including GLM-5V-Turbo, which is also frontier and closed-source.  
We also note that human performance is meaningfully above chance across all tasks, which we believe is the clearest evidence against the benchmark being ill-posed. Human accuracy on VFP (80\%) and CMD (80\%) is substantially higher than chance, indicating those tasks are well within human capability. We note here the general difficulty of the TFR task. It is fine-grained, in that one must distinguish similar frames from each other in the temporal sequence. We observe that humans perform poorly here as well, although overall performance is still meaningfully above chance. Further, we clarify that the use of shuffled and reversed frame orderings is a well-established convention in video understanding benchmarks, and notably, prior work adopting these designs has not always included human studies: for instance, AoT~\cite{xue2026seeing}, RTime~\cite{du2024reversed}, employ shuffled and reversed video evaluation without a human baseline. We believe further efforts are required to standardize evaluation comparisons between humans and frontier models, and consider this an exciting direction for future work. 

\begin{table}[t]
    \centering
    \small
    \tabcolsep 20pt %% control the table width
    \caption{\small Accuracy (\%) across the three tasks. Best result per column in \textbf{bold}. The causal memory distortion task is binary (``yes''/``no''), hence the 50\% random baseline. The other two tasks have four choices each, hence the 25\% random baseline.}
    \vspace{2mm}
    \footnotesize
    \begin{tabular}{lccc}
        \toprule
        \multirow{2}{*}{\textbf{Model}} &
        \textbf{Temporal Frame} &
        \textbf{Future} &
        \textbf{Causal Memory} \\
        & \textbf{Retrieval} & \textbf{Prediction} & \textbf{Distortion} \\
        \midrule
        \textcolor{gray}{Random Baseline}  & \textcolor{gray}{25.00} & \textcolor{gray}{25.00} & \textcolor{gray}{50.00} \\
        \textcolor{gray}{Human}            & \textcolor{gray}{44.00} & \textcolor{gray}{80.00} & \textcolor{gray}{80.00} \\
        \midrule
        LLaVA-OV-72B     & 23.62 & 25.75 & 52.00 \\[0.5ex]
        Qwen2.5-VL-32B      & 24.71 & 33.44 & 55.00 \\[0.5ex]
        Qwen2.5-VL-72B      & 25.83 & 31.70 & \textbf{60.00} \\[0.5ex]
        mPLUG-Owl3       & 17.69 & 17.40 & 51.70 \\[0.5ex]
        Qwen3-VL-32B        & 21.57 & 41.80 & 50.00 \\[0.5ex]
        InternVL3-78B    & 21.48 & 25.80 & 50.00 \\[0.5ex]
        GLM-4.6V         & 29.78 & 32.44 & 51.70 \\[0.5ex]
        Gemini-2.5-Pro   & 30.00 & \textbf{69.62} & 58.00 \\[0.5ex]
        Kimi-K2.6        & \textbf{33.45} & 28.43 & 58.33 \\[0.5ex]
        GLM-5V-Turbo     & 23.87 & 24.08 & 55.00 \\
        \bottomrule
    \end{tabular}
    \label{tab:main_results}
    % \vspace{-3mm}
\end{table}

Finally, on the causal distortion task, we present a further analysis of statistical significance and the sensitivity-specificity rates to help in the interpretation of the results, owing to comparatively lower sample size. These results are in appendix \ref{qwen-variance} and \ref{sens-spec}. We observe the rather interesting result that the specificity rate of Gemini is 30 points lower than Qwen2.5-72B, suggesting that it is much more likely to predict ``yes'', i.e., it hallucinated the contact frame at a much higher rate than Qwen.

% We leave further substantiating results on statistical significance, sensitivity/specificity, and model uncertainties in \cref{qwen-variance}, \cref{sens-spec}, \cref{perplexity} respectively.

% \begin{insight}
% All models perform slightly above the random baselines in all the tasks, except Gemini on Shapestacks. These results suggest a weak ability of frontier video models to visually ground their decision-making.
% \end{insight}

% \begin{figure}[t!]
%     \centering
%     \resizebox{0.5\linewidth}{!}{\input{figures/causal_rates}}
%     \caption{Causal evaluation results across models. Higher is better for both metrics. Gemini's higher sensitivity owes itself to its high hit rate, while its low specificity owes itself to its high false alarm rate. Overall, Qwen72B provides a more balanced performance on this task. Gemini has a propensity of saying ``yes''.}
%     \label{fig:causal_rates}
%     \vspace{-3mm}
% \end{figure}

\section{Analysis}

The broad theme of our paper is to uncover the tension between two modalities towards the evaluation of frontier video models. We seek to elaborate on our main findings with key experiments further shedding insight on the interplay between the text and vision modalities. 

\subsection{Comparison with text-based evaluation}
We begin with the most natural question: \textit{given the poor performance of models on our visual-centric tasks, how well do models perform if the evaluation setup is text-based, which is the current approach taken towards benchmark design?}
We conduct this analysis along two axes. First, we use the temporal frame retrieval task built on Charades, which provides time-stamped action descriptions that can be used directly as text options\footnote{See \cref{text-negatives} in the appendix for details on construction of the text options}. This tests whether there is a meaningful difference if text options are used as-is. Second, to test whether the gap persists when visual content is converted to text rather than drawn from pre-existing annotations, we use GPT-5.4 to caption the visual options from the Shapestacks task and evaluate Gemini on this distilled text-based version of the dataset. This tests whether there is a meaningful difference if visual data is first \textit{converted} into text, and then evaluated. 

As illustrated in \Cref{fig:text_vs_video_accuracies}, we observe a significant drop in performance across all models when the evaluation modality shifts from text to visual on the Charades task. However, in the distillation experiment on Shapestacks (\Cref{fig:text_distilled})  accuracy drops from $69.62$ with visual options to $45.12$ when those options are distilled into text captions. This loss of fine-grained visual detail through textual distillation is consistent with prior observations ~\cite{schroditwo}. Taken together, we see that text-based evaluations may inflate performance when text options are available as-is, but the \textit{opposite} may happen if visual options are converted into text, owing to loss in fine-grained information. These results demonstrate the benefit of our visual benchmark as a complementary alternative to text-based evaluations.

\begin{figure}[t!]
    \centering
    \begin{minipage}[t]{0.48\linewidth}
        \centering
        \vspace{0pt}
        \resizebox{\linewidth}{!}{\begin{tikzpicture}
\begin{axis}[
    ybar,
    bar width=0.5cm,
    width=10cm,
    height=5cm,
    ylabel={Accuracy (\%)},
    ylabel style={font=\small},
    symbolic x coords={Gemini2.5-Pro},
    xtick=data,
    xticklabel style={font=\scriptsize},
    ymin=0,
    ymax=80,
    ytick={0,20,40,60,80},
    yticklabel style={font=\scriptsize},
    legend style={
        at={(0.5,-0.2)},
        anchor=north,
        legend columns=3,
        column sep=1mm,
        inner ysep=1mm,
        inner xsep=2mm,
        font=\footnotesize,
        nodes={text height=1mm, text depth=0mm}
    },
    ymajorgrids=false,
    enlarge x limits=0.2,
    clip=false,
    clip mode=individual,
]
% Text Accuracy (Charades) - now with text values
\addplot[fill=green!20, draw=black] coordinates {
    % (LLaVA-OV-72B, 64.35)
    % (Qwen2.5-32B, 56.86)
    % (Qwen2.5-72B, 60.74)
    (Gemini2.5-Pro, 74)
};
% Visual Accuracy - now with visual values
\addplot[fill=orange!20, draw=black] coordinates {
    % (LLaVA-OV-72B, 23.62)
    % (Qwen2.5-32B, 24.71)
    % (Qwen2.5-72B, 25.83)
    (Gemini2.5-Pro, 30.0)
};
% Delta interval lines
% \draw[|-|, thick, red, xshift=0.3cm] (axis cs:LLaVA-OV-72B,64.35) -- (axis cs:LLaVA-OV-72B,23.62) node[midway, right, xshift=0.1cm, red] {$\Delta$};
% \draw[|-|, thick, red, xshift=0.3cm] (axis cs:Qwen2.5-32B,56.86) -- (axis cs:Qwen2.5-32B,24.71) node[midway, right, xshift=0.1cm, red] {$\Delta$};
% \draw[|-|, thick, red, xshift=0.3cm] (axis cs:Qwen2.5-72B,60.74) -- (axis cs:Qwen2.5-72B,25.83) node[midway, right, xshift=0.1cm, red] {$\Delta$};
\draw[|-|, thick, red, xshift=0.3cm] (axis cs:Gemini2.5-Pro,74) -- (axis cs:Gemini2.5-Pro,30.0) node[midway, right, xshift=0.1cm, red] {$\Delta$};
% 25% Random Baseline
\draw[thick, dashed, color=black!60] ([yshift=0cm]current axis.west|-{axis cs:Gemini2.5-Pro,25}) -- ([yshift=0cm]current axis.east|-{axis cs:Gemini2.5-Pro,25});
\addlegendimage{thick, dash pattern=on 1.5pt off 1pt, color=black!60}
\legend{Text, Visual, Random}
\end{axis}
\end{tikzpicture}}
        \captionof{figure}{\small Comparison of model accuracies on the Temporal Frame Retrieval task when evaluated with text options vs. video options. The significant drop in performance (large \textcolor{red}{$\Delta$}) when using visual options highlights the limitations of text-based evaluations.}
        \label{fig:text_vs_video_accuracies}
    \end{minipage}
    \hfill
    \begin{minipage}[t]{0.48\linewidth}
        \centering
        \vspace{0pt}
        \resizebox{\linewidth}{!}{\begin{tikzpicture}
\begin{axis}[
    ybar,
    bar width=0.5cm,
    width=10cm,
    height=5cm,
    ylabel={Accuracy (\%)},
    ylabel style={font=\small},
    symbolic x coords={Gemini2.5-Pro},
    xtick=data,
    xticklabel style={font=\scriptsize},
    ymin=0,
    ymax=80,
    ytick={0,20,40,60,80},
    yticklabel style={font=\scriptsize},
    legend style={
        at={(0.5,-0.2)},
        anchor=north,
        legend columns=3,
        column sep=1mm,
        inner ysep=1mm,
        inner xsep=2mm,
        font=\footnotesize,
        nodes={text height=1mm, text depth=0mm}
    },
    ymajorgrids=false,
    enlarge x limits=0.2,
    clip=false,
    clip mode=individual,
]
% Visual Accuracy (Shapestacks)
\addplot[fill=orange!20, draw=black] coordinates {
    (Gemini2.5-Pro, 69.62)
};
% Text [Distilled] Accuracy
\addplot[fill=blue!20, draw=black] coordinates {
    (Gemini2.5-Pro, 45.12)
};
% Delta interval line
\draw[|-|, thick, red, xshift=0.3cm] (axis cs:Gemini2.5-Pro,69.62) -- (axis cs:Gemini2.5-Pro,45.12) node[midway, right, xshift=0.1cm, red] {$\Delta$};
% 25% Random Baseline
\draw[thick, dashed, color=black!60] ([yshift=0cm]current axis.west|-{axis cs:Gemini2.5-Pro,25}) -- ([yshift=0cm]current axis.east|-{axis cs:Gemini2.5-Pro,25});
\addlegendimage{thick, dash pattern=on 1.5pt off 1pt, color=black!60}
\legend{Visual, Text [Distilled], Random}
\end{axis}
\end{tikzpicture}}
        \captionof{figure}{\small Gemini performance on the future prediction task when visual options are distilled into text captions. The substantial drop (large \textcolor{red}{$\Delta$}) indicates that captioning loses fine-grained visual information needed to disambiguate  options.}
        \label{fig:text_distilled}
    \end{minipage}
    \vspace{-3mm}
\end{figure}

\begin{figure*}[t]
  \centering
  \begin{minipage}[t]{0.48\linewidth}
    \centering
    \resizebox{\linewidth}{!}{% ============================================================================
% FIGURE 1: Gemini 2.5 Pro Reasoning Pattern - Option Grounding Failure
% ============================================================================

% Define colors
\definecolor{correctgreen}{RGB}{102, 170, 120}
\definecolor{errorred}{RGB}{190, 90, 85}
\definecolor{errororange}{RGB}{255, 152, 0}
\definecolor{mutedgray}{RGB}{158, 158, 158}
\definecolor{lightgray}{RGB}{240, 240, 240}
\definecolor{darkgray}{RGB}{66, 66, 66}
\definecolor{framebg}{RGB}{250, 250, 250}
\definecolor{fadedgray}{RGB}{200, 200, 200}

\begin{tikzpicture}[
    font=\sffamily\footnotesize,
    >=To,
    % Styles
    frame/.style={
        rectangle, 
        draw=mutedgray, 
        fill=framebg, 
        minimum width=0.5cm, 
        minimum height=0.4cm
    },
    eventbox/.style={
        rectangle, 
        rounded corners=3pt, 
        draw=correctgreen, 
        line width=1.5pt,
        fill=white, 
        minimum width=2.5cm, 
        minimum height=0.55cm,
        align=center,
        inner sep=4pt
    },
    modelbox/.style={
        rectangle, 
        rounded corners=5pt, 
        draw=darkgray, 
        fill=lightgray, 
        minimum width=2cm, 
        minimum height=0.8cm,
        font=\sffamily\small 
    },
    answerbox/.style={
        rectangle, 
        rounded corners=5pt, 
        draw=errorred, 
        line width=1.5pt,
        fill=white, 
        minimum width=1.8cm, 
        minimum height=0.7cm,
        font=\sffamily\small
    },
    optionlabel/.style={
        font=\sffamily\small\bfseries,
        text=darkgray
    },
    sectionlabel/.style={
        font=\sffamily\small,
        text=darkgray
    },
    arrow/.style={
        ->,
        line width=0.8pt,
        darkgray!70,
        line cap=round
    },
    bifurcatearrow/.style={
        ->,
        line width=0.8pt,
        darkgray!70,
        line cap=round
    }
]

% Force identical bounding box with reasoning_qwen.tex so resizebox scales both equally
\useasboundingbox (4.5, -1.3) rectangle (17, 3.0);

% ============================================================================
% Source Video (Filmstrip) — centered above the model
% ============================================================================
\node[sectionlabel] at (5.5, 2.4) {Source Video};

\foreach \i in {0,1,2,3,4} {
    \node[frame] (srcframe\i) at (4.3+\i*0.6, 1.75) {};
}
% Filmstrip border
\node[draw=black, rounded corners=2pt, fit=(srcframe0)(srcframe4), inner sep=2pt] (filmstrip) {};

% ============================================================================
% Column 2: Model Icon
% ============================================================================
\node[modelbox] (model) at (5.5, 0.5) {Gemini 2.5 Pro};

% ============================================================================
% Column 3: Event List
% ============================================================================
\node[sectionlabel] at (9.0, 2.5) {Extracted Events};

% Event boxes with green border
\node[eventbox] (event1) at (9.0, 1.6) {};
\node[eventbox] (event2) at (9.0, 0.9) {};
\node[eventbox] (event3) at (9.0, 0.2) {};
\node[eventbox] (event4) at (9.0, -0.5) {};

% Group box for event list (for arrow anchoring)
\node[fit=(event1)(event4), inner sep=3pt] (eventgroup) {};

% Border around extracted events
\node[draw=black, rounded corners=5pt, fit=(event1)(event4), inner sep=8pt] (eventborder) {};

% Event text
\node[anchor=center, font=\sffamily\scriptsize] at (event1.center) {Event};
\node[anchor=center, font=\sffamily\scriptsize] at (event2.center) {Event};
\node[anchor=center, font=\sffamily\scriptsize] at (event3.center) {Event};
\node[anchor=center, font=\sffamily\scriptsize] at (event4.center) {Event};

% ============================================================================
% Arrows (even spacing, straight lines)
% ============================================================================
% Arrow from filmstrip to model
\draw[arrow] (filmstrip.south) -- (model.north);

% Arrow from model to events (use calc to keep it horizontal)
\draw[arrow] (model.east) -- (model.east -| eventborder.west);

% ============================================================================
% Column 4: Visual Options (A, B, C, D filmstrips) - NOT FADED
% ============================================================================
\node[sectionlabel] at (13, 2.3) {Visual Options};

% Border around all visual options (defined first for label positioning)
\node[draw=black, rounded corners=5pt, minimum width=3.2cm, minimum height=2.8cm] (optionsborder) at (13.0, 0.55) {};

% Option A
\foreach \i in {0,1,2,3,4} {
    \node[frame, minimum width=0.4cm, minimum height=0.3cm] (optA\i) at (12+\i*0.5, 1.6) {};
}
\node[draw=mutedgray, rounded corners=2pt, fit=(optA0)(optA4), inner sep=1pt] (optAstrip) {};

% Option B - correct answer (green border)
\foreach \i in {0,1,2,3,4} {
    \node[frame, minimum width=0.4cm, minimum height=0.3cm] (optB\i) at (12+\i*0.5, 0.9) {};
}
\node[draw=correctgreen, line width=1.5pt, rounded corners=2pt, fit=(optB0)(optB4), inner sep=1pt] (optBstrip) {};

% Option C - model's wrong answer (red border)
\foreach \i in {0,1,2,3,4} {
    \node[frame, minimum width=0.4cm, minimum height=0.3cm] (optC\i) at (12+\i*0.5, 0.2) {};
}
\node[draw=errorred, line width=1.5pt, rounded corners=2pt, fit=(optC0)(optC4), inner sep=1pt] (optCstrip) {};

% Option D
\foreach \i in {0,1,2,3,4} {
    \node[frame, minimum width=0.4cm, minimum height=0.3cm] (optD\i) at (12+\i*0.5, -0.5) {};
}
\node[draw=mutedgray, rounded corners=2pt, fit=(optD0)(optD4), inner sep=1pt] (optDstrip) {};

% Option labels outside the border box (to the left)
\node[optionlabel, anchor=east] at ([xshift=0.05cm]optionsborder.west |- optAstrip) {A};
\node[optionlabel, anchor=east] at ([xshift=0.05cm]optionsborder.west |- optBstrip) {B};
\node[optionlabel, anchor=east] at ([xshift=0.05cm]optionsborder.west |- optCstrip) {C};
\node[optionlabel, anchor=east] at ([xshift=0.05cm]optionsborder.west |- optDstrip) {D};

% ============================================================================
% Bifurcating Arrow (KEY ELEMENT)
% ============================================================================
% Main arrow from events border, then bifurcates to B and C
\coordinate (bifurcatepoint) at ($(eventborder.east)!0.5!(optionsborder.west)$);

% Arrow from events to bifurcation point
\draw[bifurcatearrow, -] (eventborder.east) -- (bifurcatepoint);

% Branch to option B (correct)
\draw[bifurcatearrow] (bifurcatepoint) to[out=20, in=180] (optBstrip.west);

% Branch to option C (model's choice)
\draw[bifurcatearrow] (bifurcatepoint) to[out=-20, in=180] (optCstrip.west);

% Question mark between option B and C
\node[font=\sffamily\small\bfseries, text=darkgray] at ($(optBstrip.west)!0.5!(optCstrip.west) + (1.2cm, 0)$) {?};

% ============================================================================
% Column 5: Model Answer
% ============================================================================
\node[answerbox] (answer) at (16.2, 0.55) {Answer: C};

% Arrow from options to answer (straight, horizontal)
\draw[arrow] (optionsborder.east) -- (optionsborder.east -| answer.west);

% ============================================================================
% Caption
% ============================================================================

\end{tikzpicture}}
    \caption{\small We analyze Gemini2.5-Pro's reasoning traces and find strong event understanding for video clips, grounded in both the context video and visual options. However, it often confounds the final answer: in the figure above, it correctly identifies option B in its reasoning, yet outputs option C. 
    % This \textit{option (mis)grounding} behaviour was not replicated in Qwen2.5-72B.
    }
    \label{fig:gemini_reasoning}
  \end{minipage}\hfill
  \begin{minipage}[t]{0.48\linewidth}
    \centering
    \resizebox{\linewidth}{!}{\definecolor{correctgreen}{RGB}{102, 170, 120}
\definecolor{errorred}{RGB}{190, 90, 85}
\definecolor{mutedgray}{RGB}{158, 158, 158}
\definecolor{lightgray}{RGB}{240, 240, 240}
\definecolor{darkgray}{RGB}{66, 66, 66}
\definecolor{framebg}{RGB}{250, 250, 250}
\definecolor{fadedgray}{RGB}{200, 200, 200}

\begin{tikzpicture}[
    >=To,
    % Styles
    frame/.style={
        rectangle, 
        draw=mutedgray, 
        fill=framebg, 
        minimum width=0.5cm, 
        minimum height=0.4cm
    },
    fadedframe/.style={
        rectangle, 
        draw=fadedgray, 
        fill=white, 
        minimum width=0.4cm, 
        minimum height=0.3cm,
        opacity=0.4
    },
    eventbox/.style={
        rectangle, 
        rounded corners=3pt, 
        draw=correctgreen, 
        line width=1.5pt,
        fill=white, 
        minimum width=2.5cm, 
        minimum height=0.55cm,
        align=center,
        inner sep=4pt
    },
    modelbox/.style={
        rectangle, 
        rounded corners=5pt, 
        draw=darkgray, 
        fill=lightgray, 
        minimum width=2cm, 
        minimum height=0.8cm,
        font=\sffamily\small 
    },
    answerbox/.style={
        rectangle, 
        rounded corners=5pt, 
        draw=errorred, 
        line width=1.5pt,
        fill=white, 
        minimum width=1.8cm, 
        minimum height=0.7cm,
        font=\sffamily\small
    },
    optionlabel/.style={
        font=\sffamily\small\bfseries,
        text=fadedgray
    },
    sectionlabel/.style={
        font=\sffamily\small,
        text=darkgray
    },
    arrow/.style={
        ->,
        line width=0.8pt,
        darkgray!70,
        line cap=round
    },
    bypassarrow/.style={
        ->,
        line width=0.8pt,
        darkgray!70,
        dashed,
        dash pattern=on 2pt off 3pt,
        line cap=round
    }
]

% Force identical bounding box with reasoning.tex so resizebox scales both equally
\useasboundingbox (4.5, -1.3) rectangle (17, 3.0);

% ============================================================================
% Source Video (Filmstrip) — centered above the model
% ============================================================================
\node[sectionlabel] at (5.5, 2.4) {Source Video};

\foreach \i in {0,1,2,3,4} {
    \node[frame] (srcframe\i) at (4.3+\i*0.6, 1.75) {};
}
% Filmstrip border
\node[draw=black, rounded corners=2pt, fit=(srcframe0)(srcframe4), inner sep=2pt] (filmstrip) {};

% ============================================================================
% Column 2: Model Icon
% ============================================================================
\node[modelbox] (model) at (5.5, 0.5) {Qwen2.5-72B};

% ============================================================================
% Column 3: Event List
% ============================================================================
\node[sectionlabel] at (9.0, 2.5) {Extracted Events};

% Event boxes with green border
\node[eventbox] (event1) at (9.0, 1.6) {};
\node[eventbox] (event2) at (9.0, 0.9) {};
\node[eventbox] (event3) at (9.0, 0.2) {};
\node[eventbox] (event4) at (9.0, -0.5) {};

% Group box for event list (for arrow anchoring)
\node[fit=(event1)(event4), inner sep=3pt] (eventgroup) {};

% Border around extracted events
\node[draw=black, rounded corners=5pt, fit=(event1)(event4), inner sep=8pt] (eventborder) {};

% Event text
\node[anchor=center, font=\sffamily\scriptsize] at (event1.center) {Event};
\node[anchor=center, font=\sffamily\scriptsize] at (event2.center) {Event};
\node[anchor=center, font=\sffamily\scriptsize] at (event3.center) {Event};
\node[anchor=center, font=\sffamily\scriptsize] at (event4.center) {Event};

% ============================================================================
% Arrows (even spacing, straight lines)
% ============================================================================
% Arrow from filmstrip to model
\draw[arrow] (filmstrip.south) -- (model.north);

% Arrow from model to events (use calc to keep it horizontal)
\draw[arrow] (model.east) -- (model.east -| eventborder.west);

% ============================================================================
% Column 4: Visual Options (A, B, C, D filmstrips) - GRAYED OUT
% ============================================================================
\node[sectionlabel, text=fadedgray] at (12.75, 2.5) {Visual Options};

% Option A - faded
\node[optionlabel] at (11.2, 1.6) {A};
\foreach \i in {0,1,2,3,4} {
    \node[fadedframe] (optA\i) at (11.8+\i*0.5, 1.6) {};
}
\node[draw=fadedgray, rounded corners=2pt, fit=(optA0)(optA4), inner sep=1pt, opacity=0.4] (optAstrip) {};

% Option B - PLACEHOLDER_CORRECT_OPTION (faded green border)
\node[optionlabel] at (11.2, 0.9) {B};
\foreach \i in {0,1,2,3,4} {
    \node[fadedframe] (optB\i) at (11.8+\i*0.5, 0.9) {};
}
% Faded green border for correct answer - PLACEHOLDER: Change to correct option
\node[draw=correctgreen, line width=1.5pt, rounded corners=2pt, fit=(optB0)(optB4), inner sep=1pt, opacity=0.35] (optBstrip) {};

% Option C - faded
\node[optionlabel] at (11.2, 0.2) {C};
\foreach \i in {0,1,2,3,4} {
    \node[fadedframe] (optC\i) at (11.8+\i*0.5, 0.2) {};
}
\node[draw=fadedgray, rounded corners=2pt, fit=(optC0)(optC4), inner sep=1pt, opacity=0.4] (optCstrip) {};

% Option D - faded
\node[optionlabel] at (11.2, -0.5) {D};
\foreach \i in {0,1,2,3,4} {
    \node[fadedframe] (optD\i) at (11.8+\i*0.5, -0.5) {};
}
\node[draw=fadedgray, rounded corners=2pt, fit=(optD0)(optD4), inner sep=1pt, opacity=0.4] (optDstrip) {};

% Group box around all options (faded)
\node[draw=fadedgray, dashed, rounded corners=5pt, fit=(optAstrip)(optDstrip), inner sep=5pt, opacity=0.5] (optionsgroup) {};

% ============================================================================
% Column 5: Model Answer
% ============================================================================
% PLACEHOLDER: Change C to model's answer
\node[answerbox] (answer) at (16.0, 0.625) {Answer: C};

% Red X
\node[text=errorred, font=\large\bfseries] at ([xshift=0.5cm]answer.east) {};

% ============================================================================
% Bypass Arrow (KEY ELEMENT)
% ============================================================================
% Dashed curved arrow that bypasses the options entirely (top arrow)
\draw[bypassarrow] 
    ([xshift=0.1cm, yshift=0.2cm]event1.east) 
    to[out=30, in=150] 
    ([yshift=0.1cm]answer.west);

% Label between the two bypass arrows
\node[font=\sffamily\tiny\itshape, text=darkgray, opacity=0.7] at (12.8, 0.5) {No option analysis};

% Secondary bypass indicator arrow (bottom arrow)
\draw[bypassarrow, opacity=0.6] 
    ([xshift=0.1cm]event4.east) 
    to[out=-40, in=-150] 
    ([yshift=-0.1cm]answer.west);

% ============================================================================
% Caption
% ============================================================================

\end{tikzpicture}}
    \caption{\small We analyze Qwen2.5-72B's reasoning trace and find strong event understanding for the context video (it identifies key actions), but it fails to ground its decision in the visual options. In the illustration, it skips analyzing the options entirely and selects option C, when the correct answer was B. 
    % This \textit{option skipping} behaviour was not replicated in Gemini2.5-Pro.
    }
    \label{fig:qwen_reasoning}
  \end{minipage}
  \label{fig:race}
  % \vspace{-3mm}
\end{figure*}

\subsection{Reasoning Patterns}

We manually evaluate reasoning traces from Gemini-2.5 Pro and Qwen2.5-72B on the temporal frame retrieval task by extracting and annotating claims for factual grounding. The two models show distinct reasoning patterns. Gemini first describes the video broadly, then analyzes options one by one. While it correctly identifies semantic content and temporal structure, it frequently misidentifies reversed sequences and fails to map its reasoning to the correct visual option---a failure of \textit{option grounding}, illustrated in \Cref{fig:gemini_reasoning}. Qwen, by contrast, ignores the visual options entirely, listing discrete events before selecting an answer. It correctly identifies events in $\sim$66\% of cases, but without grounding to the options, performance suffers---a failure of \textit{query grounding}, illustrated in \Cref{fig:qwen_reasoning}. Both patterns illustrate our core motivation: models reason well on text but struggle to ground reasoning in the visual modality. Addressing option grounding in Gemini and query grounding in Qwen are natural future directions, though prompting strategies remain unexplored. We present sample traces and other details on reasoning and error patterns in \cref{reasoning-analysis}, and \cref{error-analysis} respectively.

\subsection{Harder text options}
\label{text-negatives}
For the text options, scrambled/reversed descriptions are sampled from different videos in the corpus to avoid confounders, since these sequences are not real actions in the video. The correct option and closest distractor use exact Charades timestamp descriptions. To ensure that this design does not create easy negatives, we use VideoCon \cite{bansal2024videocon}, to create harder textual negatives and evaluate Gemini on it as a sanity check against artificial inflation of performance. The goal is to ensure that the distractor modality, rather than its difficulty, is the key issue. As shown in \cref{tab:modality_accuracy}, accuracy drops only marginally from 73.87 (original text) to 71.48 (hard text), while the gap to visual querying (30.00) remains large. We believe this supports our hypothesis that visual querying uncovers aspects of fine-grained visuo-temporal understanding that text options cannot.

\begin{table}[h]
\centering
\begin{tabular}{lc}
\toprule
Modality & Acc. \\
\midrule
Text [Original] & 73.87 \\
Text [Hard] & 71.48 \\
Visual & 30.00 \\
\bottomrule
\end{tabular}
\caption{\small Accuracy for different text negatives}
\label{tab:modality_accuracy}
\end{table}

\subsection{Prompt sensitivity}
We run an ablation for Gemini and Kimi K2.6 on TFR using prompt variations. We modify the original prompt's wording and structure, but keep the semantics unchanged. If prompt sensitivity were the key differentiator, we would observe a significant difference in the vision-text performance gap. However, consistent with our central claim, we observe a minimal difference in the vision-text performance gap, demonstrated in \cref{tab:text-visual-performance}. It cannot, therefore, be attributed to variations in the prompt.

\begin{table}[h]
\centering
\caption{\small Text and visual performance across model variants. $\Delta$ denotes the difference between visual and text performance.}
\footnotesize
\begin{tabular}{lccc}
\toprule
Variant & Text & Visual & $\Delta$ \\
\midrule
Gemini [Original] & 0.739 & 0.307 & $-0.432$ \\
Gemini [Modified] & 0.744 & 0.306 & $-0.438$ \\
Kimi [Original]   & 0.721 & 0.334 & $-0.386$ \\
Kimi [Modified]   & 0.714 & 0.256 & $-0.458$ \\
\bottomrule
\end{tabular}
\label{tab:text-visual-performance}
\end{table}

This is consistent with our central claim in the paper - visual querying surfaces gaps that text querying alone cannot elucidate in video understanding.

\section{Limitations}
First, we wish to further expand the dataset
we created. This will not be trivial: in addition to the manual curation required for the causal memory distortion task, picking appropriate datasets for the other two tasks would require careful deliberation as well. Second, our analysis in this paper was
entirely black-box, i.e. it does not include a key complementary view of our paper: that of model representations and internal mechanisms. In the future, we would like to explore white-box perspectives on this work, particularly the role of the vision encoder in inference.
\vspace{-2.5mm}
\section{Conclusion}
In this work, we selected the domain of video understanding in VQA tasks and introduced a new evaluation approach for frontier video models through visual queries. We introduced three fundamental tasks: temporal frame retrieval, video future prediction, and causal memory distortion. Our results shed light on the role that visual information plays as an evaluation modality for frontier video models, as opposed to text, and demonstrated that these models still have significant weaknesses in visual understanding. We believe that our work provides a complementary view to existing text-based benchmarks. It does not replace them, but rather seeks to complement them to improve our understanding of how these models operate. Finally, we hope that this new evaluation approach can facilitate more analyses of the language-based biases to which text-based reasoning benchmarks may be susceptible.

\section{Impact Statement}
This work advances the evaluation of video--language models with a focus on visuo-temporal understanding. We provide a diagnostic lens to understanding how frontier video models fail on direct visual tasks. Our evidence thus paves the way for more targeted model developments that reduce the reliance on modal shortcuts. We have, to the best of our abilities, ensured that the typical issues with biased and problematic content generation associated with frontier models, do not arise in our work. Another key issue to address is the heavy compute associated with running model evaluations. The total compute used in this work is 240 GPU hours on NVIDIA GH200 (96GB) and 147 CPU hours. We acknowledge the need for improving efficient inference pipelines in the future. 

\section{Funding statement}
This work was supported by the Research Council of Norway (NFR), through its Centre for Research-based Innovation (grant no. 309439).

% Perhaps "true" understanding stems
% from a sense of consistency - if models are correct, they
% should be correct and robust across diverse modalities.

% \begin{ack}
% Use unnumbered first level headings for the acknowledgments. All acknowledgments

% Do {\bf not} include this section in the anonymized submission, only in the final paper. You can use the \texttt{ack} environment provided in the style file to automatically hide this section in the anonymized submission.
% \end{ack}

\bibliographystyle{plain}
\bibliography{example_paper}

%%%%%%%%%%%%%%%%%%%%%%%%%%%%%%%%%%%%%%%%%%%%%%%%%%%%%%%%%%%%

\newpage
\appendix
\section*{Appendix}
% \section*{Impact Statement}
% This work advances the evaluation of video--language models with a focus on visuo-temporal understanding. We provide a diagnostic lens to understanding how frontier video models fail on direct visual tasks. Our evidence thus paves the way for more targeted model developments that reduce the reliance on modal shortcuts. We have, to the best of our abilities, ensured that the typical issues with biased and problematic content generation associated with frontier models, do not arise in our work. Another key issue to address is the heavy compute associated with running model evaluations. The total compute used in this work is 240 GPU hours on NVIDIA GH200 (96GB) and 147 CPU hours. We acknowledge the need for improving efficient inference pipelines in the future. 

\section{The Dataset}
\label{app:dataset}
In this section we furnish some additional details on the construction of the temporal retrieval task, the future prediction task, and the causal memory distortion task.

\subsection{Statistics}
In \cref{tab:diff_datasets}, we present detailed statistics on our benchmarks, including the number of videos, questions, frames, duration, and difficulty based on human judgement. Note, the frames column describes the total number of frames the model sees in the input query over the whole data. 
\begin{table}[ht]
\centering
\begin{tabular}{lccccc}
\hline
\textbf{Dataset} & \textbf{Videos} & \textbf{Questions} & \textbf{Frames} & \textbf{Duration} & \textbf{Difficulty} \\
\hline
Charades    & 321 & 1{,}108 & 35{,}000 & $\sim$30s & Hard     \\
Shapestacks & 300 & 300     & 12{,}000 & 1--3s     & Easy     \\
Causal      & 20  & 60      & 20       & $\sim$30s & Moderate \\
\hline
\end{tabular}
\caption{Dataset statistics.}
\label{tab:diff_datasets}
\end{table}

\subsection{Temporal Frame Retrieval}
There are 157 action classes in the Charades dataset, with timestamped annotations for each video. The annotations look like so:
\begin{lstlisting}
c077 12.10 18.00;c079 11.80 17.30;c080 13.00 18.00;
c076 11.80 17.50;c075 5.40 14.10
\end{lstlisting}
Given these annotations, we can map them to actual actions using the \texttt{\_classes.txt} file, where the text descriptions of these action ids reside. 

\subsubsection{Preprocessing}
We filter actions to keep only those with durations between 1 and 5 seconds.
    
\begin{enumerate}
    \item Actions shorter than 1 s or longer than 5 s are dropped.
    \item Videos themselves are never removed.
    \item We set thresholds of less than 1s to remove micro-actions, and more than 5s to remove general actions that are true overall but not for specific contexts. For example, in 
\begin{lstlisting}
c015 0.00 32.10;c087 0.60 32.10;c016 0.00 32.10;c154 0.00 31.8
\end{lstlisting}
both \texttt{c015} (Holding a phone), and \texttt{c154} (Standing up somewhere) are included, which are not specific enough as actions. In fact, all four actions annotated in this video are true for the whole video. We keep only discrete actions in their respective temporal windows, and not general actions that are true for the whole video. 
\end{enumerate}

We present an example JSON entry from the data:

\begin{lstlisting}
    "0HGNK": [
    {
      "id": "0HGNK_0000",
      "video_id": "0HGNK",
      "type": "after",
      "question": "What happens after the following action: Someone is going from standing to sitting?",
      "options": {
        "A": {
          "frames": [],
          "span": [
            10.3,
            15.2
          ],
          "class_id": "c152",
          "text_option": "Opening a book",
          "label": "incorrect",
          "reverse_order": true,
          "distractor_type": "reversed"
        },
        "B": {
          "frames": [],
          "span": [
            18.376751109861242,
            22.95627907087428
          ],
          "class_id": "c150",
          "text_option": "Someone is running somewhere",
          "label": "incorrect",
          "distractor_type": "random",
          "scrambled": true
        },
        "C": {
          "frames": [],
          "span": [
            0.0,
            3.8
          ],
          "class_id": "c151",
          "text_option": "Someone is going from standing to sitting",
          "label": "incorrect",
          "distractor_type": "neighbor"
        },
        "D": {
          "frames": [],
          "span": [
            10.3,
            15.2
          ],
          "class_id": "c152",
          "text_option": "Someone is smiling",
          "label": "correct"
        }
      },
      "metadata": {
        "context_class_id": "c151",
        "correct_class_id": "c152",
        "context_span": [
          0.0,
          3.8
        ],
        "correct_span": [
          10.3,
          15.2
        ],
        "correct_option": "D"
      }
\end{lstlisting}
All the necessary information required to load the correct options are apparent in this structure. We will release all the JSON files for the three tasks.

\subsection{Video Future Prediction}

Shapestacks is a 3D simulation dataset that creates RGB renderings of geometrical primitives (e.g. cube, cuboids, cylinders, spheres) under two structurally unstable configurations - a center-of-mass violation, and a planar surface violation.

Each configuration is of the form: \begin{lstlisting}
env_<subset_name>-<height>-<vcom>-<vpsf>-<version>
\end{lstlisting}

where:
\begin{enumerate}
    \item \texttt{subset\_name}: Indicates whether the configuration contains only cubes (blocks) or cuboids, cylinders and spheres (ccs).
    \item height h: the number of objects in the stack
    \item vcom: the layer of the stack in which the stability violation occurs. Layers are counted from bottom (=1) to top (=6).
    \item vpsf: annotates the layer of the stack in which a planar surface violation occurs. Layers are counted from bottom (=1) to top (=6).
    \item version v: the version number of this scenario, since scenarios with the same parameters have been sampled multiple times.
\end{enumerate}

There are both \textit{easy} and \textit{hard} versions of these configurations. In our dataset, we sample in a balanced manner across both the \textit{blocks} and \textit{ccs} types, and the \textit{easy} and \textit{hard} configurations. We truncate each video to eight \textit{input} frames, and the remaining eight as \textit{label} frames. Then, we create the four candidate options as discussed in the main paper. 

Here is an example of the JSON entry:

\begin{lstlisting}
    "env_ccs-hard-h=4-vcom=1-vpsf=0-v=98": {
    "correct_option": "C",
    "options": {
      "C": "correct",
      "A": "rev_correct",
      "B": "scrambled",
      "D": "rev_input"
    }
\end{lstlisting}

Each of the options are saved as frames in their respective subdirectories under the main directory (here, \texttt{env\_ccs-hard-h=4-vcom=1-vpsf=0-v=98}). In the end, we have 300 videos. 

\subsection{Causal Memory Distortion}

Since the original study by \cite{Strickland2011EventCE} did not release the videos, we created our own annotation tool, and used the Charades dataset as the base to create 20 clips, following the design protocol in the original study. We do not include \textit{fillers} or \textit{lures} as these were added as additional controls for human distractions, which we believe do not readily apply to the model case. In our annotation tool, we manually select the temporal window where the contact (e.g. throwing a ball) occurs, and make sure to remove successive frames from the contact frame, as the task is \textit{not} fine-grained frame retrieval. Rather, we want a set of frames following the contact frame that are either causally indicative of contact, or not. For example, if a person is throwing a ball, we do not need all the frames capturing the ball's trajectory. We need the key frame, i.e. the moment where the ball leaves the hand (contact), and the following frames as the ball in the air/on the ground. This is a causal implication, and is thus not strictly a fine-grained frame retrieval task. 

Here is an example JSON entry:

\begin{lstlisting}
     {
      "video_id": "10M0F",
      "versions": {
        "complete": "causal/benchmark/videos/10M0F_complete.mp4",
        "incomplete_with_implication": "causal/benchmark/videos/10M0F_incomplete_with_implication.mp4",
        "incomplete_without_implication": "causal/benchmark/videos/10M0F_incomplete_without_implication.mp4",
        "scrambled": "causal/benchmark/videos/10M0F_scrambled.mp4"
      },
      "frames":
        {
          "frame_id": "contact",
          "frame_type": "contact",
          "path": "frames/10M0F_contact.jpg",
          "answers": {
            "complete": "yes",
            "incomplete_with_implication": "no",
            "incomplete_without_implication": "no",
            "scrambled": "no"
          }
        }
    }
\end{lstlisting}

As in the original paper, we make sure the clips are all time-matched to within 1 second of the complete video with causal implication, ensuring that the number of frames the model sees will be approximately equal. 

\section{Evaluation}
\label{app:eval}
As opposed to the standard text-based evaluation, our dataset allows us to interleave visual inputs in the prompt. Here, we discuss the task specific prompt templates used in our paper. In general, we use the \texttt{transformers} library from HuggingFace for all the open-source models, which has a convenient interface to interleave the frame options and pass it to the model. We take the same approach for the Gemini API as well. 

\subsection{Temporal Frame Retrieval}
In the temporal frame retrieval task, the example prompt looks like:

\begin{lstlisting}
    You are an expert at video understanding. You will be shown a video and multiple-choice question about events in it.\n\n"
        f"Question: {q}\n\n"
        "Options:"
        "A": <Load frames from Option A>
        "B": <Load frames from Option B>
        ...
        \n\n
        Show your reasoning steps, and then end your response with "
        "\"The final answer is: <your answer from options A/B/C/D>\"
\end{lstlisting}

Here, \texttt{q} is loaded from the JSON file presented earlier. We extract the answers and match them with the ground truths to compare the accuracy.

\subsection{Video Future Prediction}

In this setting, we load the input frames from the appropriate paths for each video id, and the prompt structure looks like so:

\begin{lstlisting}
    "You are shown an initial state of a block tower, followed by four possible outcomes (A, B, C, D).\n\nQuestion: Which option shows the most likely outcome for this tower after some time passes?\n\nConsider physics and stability when making your prediction.\n\nInitial tower state:

    <video frames>

    \n\nShow your reasoning steps, and then end your response with "
     "\"The final answer is: <your answer from options A/B/C/D>\
\end{lstlisting}

We extract the answers and match them with the ground truths to compute the accuracy.

\subsection{Causal Memory Distortion}

Since this task is not multiple choice, we use a slightly different evaluation scheme. 

As in the original paper, the complete and incomplete assignments to each model are randomized. Since there are twenty videos, we assign ten complete and ten incomplete videos to each model. Since each video type has three conditions (with implication, without implication, scrambled), we have sixty clips in total (thirty with the complete type, and thirty with the incomplete).

We first iterate through all the complete types, and then the incomplete types. In both cases, the prompt structure is the same:

\begin{lstlisting}
    "<video_frames>

    \n Did this frame appear in the video?: <contact_frame>
    \n Answer only Yes or No."
\end{lstlisting}

Note that in this task, the ground truth is simpler - for the complete video type, the ground truth is always “Yes", since by construction, the contact frame always appear. In the incomplete video type, the ground truth is always “No", since by construction, the contact frame does \textit{not} appear.

% \section{Constructing the text options}
% \label{text-negatives}
% For the text options, scrambled/reversed descriptions are sampled from different videos in the corpus to avoid confounders, since these sequences are not real actions in the video. The correct option and closest distractor use exact Charades timestamp descriptions. To ensure that this design does not create easy negatives, we use VideoCon \cite{bansal2024videocon}, to create harder textual negatives and evaluate Gemini on it as a sanity check against artificial inflation of performance. If distractor difficulty rather than modality were the primary factor, we would expect a substantial drop in performance on this harder caption set. As shown in \cref{tab:modality_accuracy}, this is not the case: accuracy drops only marginally from 73.87 (original text) to 71.48 (hard text), while the gap to visual querying (30.00) remains large. We believe this supports our hypothesis that visual querying exposes gaps in fine-grained temporal understanding that textual querying alone cannot.

% \begin{table}[h]
% \centering
% \begin{tabular}{lc}
% \toprule
% Modality & Acc. \\
% \midrule
% Text [Original] & 73.87 \\
% Text [Hard] & 71.48 \\
% Visual & 30.00 \\
% \bottomrule
% \end{tabular}
% \caption{\small Accuracy for different text negatives}
% \label{tab:modality_accuracy}
% \end{table}

\section{Reasoning Analysis}
\label{reasoning-analysis}
In this section, we outline the steps taken to generate the results in figures \ref{fig:qwen_reasoning} and \ref{fig:gemini_reasoning}. Given the reasoning trace in the model output, we first extract claims using a large language model (GPT5.2). These models are quite capable today for this simple task. We ensure to extract only the claims (facts/assertions) in the trace, and not meta-resoning, decision justification, or final answer statements. We extract only the \textit{incorrect} predictions to probe where the models went wrong. We present below an example entry from the claim JSON we create for Qwen72B on Charades:

\begin{lstlisting}
    "T5ECU": {
    "id": "T5ECU_0000",
    "predicted": "C",
    "correct": "B",
    "claims": [
      "1. Initially, there are two people standing near a window.",
      "2. One person appears to be holding a cup and drinking from it.",
      "3. The other person seems to be adjusting something on the window or interacting with the blinds.",
      "4. There is no clear indication of anyone sitting down in the provided frames."
    ]
  },
\end{lstlisting}

For Gemini, we present an example entry below:

\begin{lstlisting}
    "QQMY8": {
    "id": "QQMY8_0000",
    "predicted": "A",
    "correct": "C",
    "claims": [
      "1.  **Analyze the initial action:** The video begins in a dimly lit room. At timestamp 00:02, the person reaches up and the room becomes brightly lit. This is the action of \"Turning on a light.\"",
      "2.  **Observe the subsequent events:** Immediately after the light turns on, the person begins to remove their blue hoodie. They pull it over their head, revealing a white t-shirt underneath.",
      "3.  **Compare with Option A:** This option shows the person, in the now brightly lit room, taking off their blue hoodie. This sequence of movements directly follows the action of turning on the light.",
      "4.  **Compare with Option B:** This option shows the person after the hoodie is already off. They are holding a yellow ball and interacting with items on a dresser. This happens later in the video, after the hoodie has been removed.",
      "5.  **Compare with Option C:** This option is a reversed or jumbled version of the events shown in Option B. It does not correctly follow the initial action.",
      "6.  **Compare with Option D:** This option includes the moment the light is turned on and then shows the person taking off the hoodie. However, the question asks what happens *after* the light is turned on. Option A is a more precise answer as it depicts only the events that follow the specified action."
    ]
  },
\end{lstlisting}

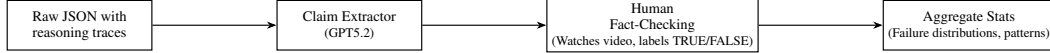
\begin{figure*}
    \centering
    \resizebox{\linewidth}{!}{\tikzset{
    process/.style={
        rectangle,
        minimum width=3.5cm,
        minimum height=1.2cm,
        text centered,
        draw=black,
        thick,
        align=center
    },
    arrow/.style={
        thick,
        -Stealth
    }
}
\begin{tikzpicture}[node distance=3cm]
% Nodes
\node (json) [process] {Raw JSON with\\reasoning traces};
\node (extractor) [process, right=of json] {Claim Extractor\\{\small (GPT5.2)}};
\node (human) [process, right=of extractor] {Human\\Fact-Checking\\{\small (Watches video, labels TRUE/FALSE)}};
\node (aggregate) [process, right=of human] {Aggregate Stats\\{\small (Failure distributions, patterns)}};
% Arrows
\draw [arrow] (json) -- (extractor);
\draw [arrow] (extractor) -- (human);
\draw [arrow] (human) -- (aggregate);
\end{tikzpicture}}
    \caption{\small The pipeline to analyze the reasoning traces in Gemini and Qwen72B}
    \label{fig:reasoning_trace}

\end{figure*}
Given these claims, we manually watch the videos and annotate which of these are true, and which are false. This allows us to evaluate how specifically the models fail at this task. The revelations of option confounding in Gemini, and option skipping in Qwen72B were thus possible using this approach. We illustrate this pipeline in Figure \ref{fig:reasoning_trace}. 

Next, we also analyzed incorrect Gemini reasoning traces on Shapestacks and identified two dominant failure patterns. Gemini correctly predicts that unstable towers will collapse in most of its responses, indicating a degree of physical reasoning. However, it frequently confuses the correct sequence to be scrambled, while assuming the opposite of the incorrect sequence (i.e., a scrambled sequence appears as coherent). This failure mode is sequence order confusion occurs in about 52.9\% of the cases, and we term it \textit{sequence order confusion}. Another dominant pattern is the tendency to perceive all options as visually identical. In these cases, it is unable to make an informed choice. This occurs in about 34.5\% of the cases, and we term it \textit{static ambiguity}. 

\section{Error Analysis}\label{error-analysis}
On temporal frame retrieval, both models' dominant error is the neighbor distractor (Gemini 37.5\%, Qwen 41.7\%) -- the temporally adjacent but wrong action -- suggesting models struggle most when fine-grained temporal discrimination is required. The reversed distractor is the least confusing (Gemini 29.4\%, Qwen 24.8\%), indicating some ability to detect temporal direction. Both models also perform substantially worse on 'before' questions (Gemini 27.0\% vs 34.6\%, Qwen 19.8\% vs 27.7\%), revealing that backward temporal reasoning is harder than forward. On Shapestacks, the models exhibit different failure modes: About 55.4\% of the time, Qwen identifies the right visual content but cannot distinguish temporal direction, whereas Gemini picks the scrambled option about 44.8\% of the time. Finally, on causal memory distortion (Appendix D.1), Gemini hallucinates contact frames at higher rates than Qwen, in addition to higher false alarm rates on average. 

\section{Additional Analysis}

\subsection{Variance on Causal Memory Distortion}
\label{qwen-variance}

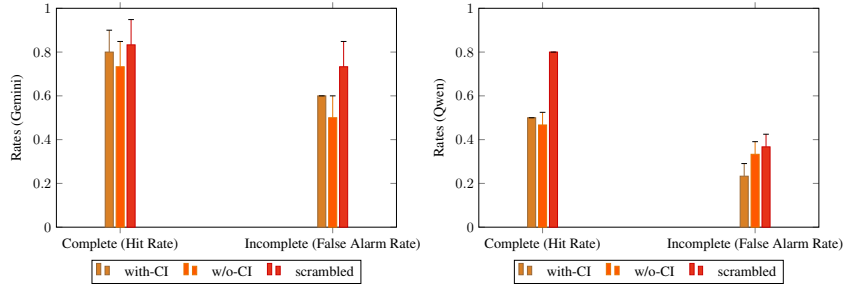
\begin{figure}
    \centering
    \resizebox{0.8\textwidth}{!}{% Gemini Plot
\begin{tikzpicture}
\begin{axis}[
    width=10cm,
    height=7cm,
    ybar,
    bar width=0.2cm,
    ylabel={Rates (Gemini)},
    title={},
    symbolic x coords={Complete (Hit Rate), Incomplete (False Alarm Rate)},
    xtick=data,
    legend style={at={(0.5,-0.15)}, anchor=north, legend columns=3, column sep = 0.3cm},
    ymin=0,
    ymax=1.0,
    enlarge x limits=0.3,
    error bars/y dir=plus,
    error bars/y explicit,
]

% with-CI condition
\addplot[fill=brown!60!orange, draw=brown!80!black] 
    coordinates {
        (Complete (Hit Rate), 0.8) +- (0, 0.1)
        (Incomplete (False Alarm Rate), 0.6) +- (0, 0)
    };
\addlegendentry{with-CI}

% w/o-CI condition
\addplot[fill=orange!70!red, draw=orange!90!black] 
    coordinates {
        (Complete (Hit Rate), 0.733) +- (0, 0.1155)
        (Incomplete (False Alarm Rate), 0.5) +- (0, 0.1)
    };
\addlegendentry{w/o-CI}

% scrambled condition
\addplot[fill=red!60!brown, draw=red!80!black] 
    coordinates {
        (Complete (Hit Rate), 0.833) +- (0, 0.1155)
        (Incomplete (False Alarm Rate), 0.733) +- (0, 0.1155)
    };
\addlegendentry{scrambled}

\end{axis}
\end{tikzpicture}

% Qwen Plot
\begin{tikzpicture}
\begin{axis}[
    width=10cm,
    height=7cm,
    ybar,
    bar width=0.2cm,
    ylabel={Rates (Qwen)},
    title={},
    symbolic x coords={Complete (Hit Rate), Incomplete (False Alarm Rate)},
    xtick=data,
    legend style={at={(0.5,-0.15)}, anchor=north, legend columns=3, column sep = 0.3cm},
    ymin=0,
    ymax=1.0,
    enlarge x limits=0.3,
    error bars/y dir=plus,
    error bars/y explicit,
]

% with-CI condition
\addplot[fill=brown!60!orange, draw=brown!80!black] 
    coordinates {
        (Complete (Hit Rate), 0.5) +- (0, 0)
        (Incomplete (False Alarm Rate), 0.233) +- (0, 0.0577)
    };
\addlegendentry{with-CI}

% w/o-CI condition
\addplot[fill=orange!70!red, draw=orange!90!black] 
    coordinates {
        (Complete (Hit Rate), 0.467) +- (0, 0.0577)
        (Incomplete (False Alarm Rate), 0.333) +- (0, 0.0577)
    };
\addlegendentry{w/o-CI}

% scrambled condition
\addplot[fill=red!60!brown, draw=red!80!black] 
    coordinates {
        (Complete (Hit Rate), 0.8) +- (0, 0)
        (Incomplete (False Alarm Rate), 0.367) +- (0, 0.0577)
    };
\addlegendentry{scrambled}

\end{axis}
\end{tikzpicture}}
    \caption{\small Average hit rates and false alarm rates on Gemini (left) and Qwen72B (right)}.
    \label{fig:qwen_average}
\end{figure}

Due to the low number of videos in the causal task (twenty), we re-ran the experiment three times to get the means and standard deviations for Qwen72B and Gemini. We did not choose the other models in this task as these two models turned out to be the best in the causal task. We visualize the average hit rates and false alarm rates on Qwen72B, in figure \ref{fig:qwen_average}.

For Gemini, we observe higher false alarm rates on average than Qwen72B, suggesting it's tendency to answer in the affirmative ("Yes") most of the time. On Qwen72B, we observe lower false alarm rates on average, explaining how the overall accuracy for this model was higher than Gemini. Overall, we find that the strong confounding effect of the incomplete condition with implication as reported in the original paper was only \textit{partially} observed in Gemini, but not observed at all in Qwen72B. In fact, in the latter model, on average the false alarm rate on the incomplete condition with causal implication was lower than without. However, we make no claims of statistical significance here, only about the means. We can conclusively conclude that Gemini does hallucinate frames more on average then Qwen72B, but we cannot conclusively conclude that \textit{within condition} differences are statistically significant.

\subsection{Sensitivity and Specificity on Causal Memory Distortion}
\label{sens-spec}

\begin{table}[h]
\centering
\caption{\small Signal detection performance across models. Sensitivity = hit rate; Specificity = $1 - \text{false alarm rate}$. $d' = z(\text{Hit}) - z(\text{FA})$ measures discriminability; $c = -\tfrac{1}{2}(z(\text{Hit}) + z(\text{FA}))$ measures response bias, where $c > 0$ indicates a conservative (``no'') bias and $c < 0$ indicates a liberal (``yes'') bias.}
\footnotesize
\begin{tabular}{lccccc}
\toprule
Model & Sensitivity (\%) & Specificity (\%) & Accuracy (\%) & $d'$ & $c$ \\
\midrule
Qwen2.5-72B     & 57 & 63 & 60 & 0.51 & $+0.08$ \\
Gemini2.5-Pro   & 83 & 33 & 58 & 0.51 & $-0.70$ \\
\midrule
Chance          & 50 & 50 & 50 & 0.00 & $\phantom{+}0.00$ \\
\bottomrule
\end{tabular}
\label{tab:signal-detection}
\end{table}

Finally, on the causal distortion task, we find the surprising result where Qwen2.5-72B outperforms Gemini 2.5 Pro slightly, suggesting that a further analysis of the sensitivity and specificity rates could help to understand this observation. The former measures how well the model performs on true positives, i.e., the model predicted ``yes'' when the ground truth was ``yes'' (the complete video type). The latter measures how well the model performs on true negatives (i.e., the model predicted ``no'' when the ground truth was ``no'').
Our results in~\Cref{tab:signal-detection} offer evidence towards why Gemini performs slightly worse - we observe that the specificity rate of Gemini is 30 points lower than Qwen, suggesting that it is much more likely to predict ``yes'', i.e., it hallucinated the contact frame at a much higher rate than Qwen. 
\subsection{Modal Perplexity}
\label{perplexity}
Models on our benchmark often behave as if they are guessing when options are videos. However, we want to ask: \textit{are models aware that they do not understand the video options (high uncertainty), or are they confidently committing to essentially random choices?} Specifically, we perform uncertainty analysis for the temporal frame retrieval task, where \textit{both} visual and text options are available. 

\begin{figure}[!t]
    \resizebox{\linewidth}{!}{\begin{tikzpicture}

% ===== GEMINI PLOT (TOP) =====
\begin{axis}[
    name=gemini,
    width=10cm,
    height=3cm,
    title={\textbf{Gemini2.5-Pro}},
    title style={font=\small, yshift=-2mm},
    xlabel={},
    ytick={1, 2},
    yticklabels={Visual, Text},
    yticklabel style={font=\small},
    tick label style={font=\footnotesize},
    xmin=0.9,
    xmax=4.2,
    ymin=0.5,
    ymax=2.5,
    xmajorgrids=true,
    ymajorgrids=false,
    grid style={line width=0.2pt, draw=gray!15},
    axis lines=left,
    axis line style={gray!70},
    tick style={gray!70},
    enlarge y limits=0.15,
    boxplot/draw direction=x,
]

% ===== GEMINI VISUAL BOXPLOT =====
\addplot+[
    boxplot prepared={
        lower whisker=1.0000,
        lower quartile=1.0018,
        median=1.0113,
        upper quartile=1.0787,
        upper whisker=1.1882,
    },
    boxplot/draw position=1,
    boxplot/box extend=0.6,
    fill=visualblue!25,
    draw=visualblue,
    thick,
] coordinates {};

% Gemini Visual outliers (182 values)
\addplot[only marks, mark=*, mark size=1.8pt, visualblue, opacity=0.2] coordinates {
    (1.195, 1) (1.197, 1) (1.198, 1) (1.198, 1) (1.203, 1) (1.207, 1) (1.208, 1) (1.208, 1) (1.208, 1) (1.209, 1)
    (1.211, 1) (1.212, 1) (1.215, 1) (1.217, 1) (1.219, 1) (1.224, 1) (1.230, 1) (1.230, 1) (1.232, 1) (1.232, 1)
    (1.234, 1) (1.234, 1) (1.234, 1) (1.234, 1) (1.238, 1) (1.239, 1) (1.240, 1) (1.243, 1) (1.247, 1) (1.247, 1)
    (1.254, 1) (1.255, 1) (1.257, 1) (1.259, 1) (1.259, 1) (1.264, 1) (1.266, 1) (1.272, 1) (1.276, 1) (1.284, 1)
    (1.286, 1) (1.295, 1) (1.300, 1) (1.310, 1) (1.311, 1) (1.313, 1) (1.313, 1) (1.315, 1) (1.317, 1) (1.319, 1)
    (1.325, 1) (1.326, 1) (1.326, 1) (1.330, 1) (1.332, 1) (1.334, 1) (1.345, 1) (1.349, 1) (1.352, 1) (1.354, 1)
    (1.358, 1) (1.360, 1) (1.362, 1) (1.364, 1) (1.378, 1) (1.383, 1) (1.388, 1) (1.389, 1) (1.392, 1) (1.401, 1)
    (1.406, 1) (1.423, 1) (1.429, 1) (1.430, 1) (1.437, 1) (1.451, 1) (1.454, 1) (1.458, 1) (1.460, 1) (1.463, 1)
    (1.468, 1) (1.470, 1) (1.480, 1) (1.482, 1) (1.508, 1) (1.512, 1) (1.524, 1) (1.529, 1) (1.538, 1) (1.557, 1)
    (1.559, 1) (1.566, 1) (1.569, 1) (1.578, 1) (1.580, 1) (1.584, 1) (1.590, 1) (1.596, 1) (1.675, 1) (1.676, 1)
    (1.680, 1) (1.695, 1) (1.702, 1) (1.720, 1) (1.740, 1) (1.766, 1) (1.780, 1) (1.802, 1) (1.805, 1) (1.805, 1)
    (1.815, 1) (1.815, 1) (1.819, 1) (1.821, 1) (1.853, 1) (1.864, 1) (1.871, 1) (1.876, 1) (1.880, 1) (1.918, 1)
    (1.930, 1) (1.932, 1) (1.933, 1) (1.934, 1) (1.948, 1) (1.948, 1) (1.949, 1) (1.949, 1) (1.951, 1) (1.956, 1)
    (1.959, 1) (1.960, 1) (1.964, 1) (1.974, 1) (1.975, 1) (1.981, 1) (1.986, 1) (1.986, 1) (1.991, 1) (1.995, 1)
    (1.995, 1) (1.998, 1) (2.000, 1) (2.000, 1) (2.003, 1) (2.008, 1) (2.014, 1) (2.014, 1) (2.016, 1) (2.016, 1)
    (2.026, 1) (2.044, 1) (2.051, 1) (2.056, 1) (2.063, 1) (2.086, 1) (2.093, 1) (2.096, 1) (2.100, 1) (2.109, 1)
    (2.118, 1) (2.174, 1) (2.178, 1) (2.188, 1) (2.274, 1) (2.359, 1) (2.377, 1) (2.433, 1) (2.438, 1) (2.450, 1)
    (2.473, 1) (2.479, 1) (2.482, 1) (2.484, 1) (2.547, 1) (2.573, 1) (2.583, 1) (2.652, 1) (2.796, 1) (2.846, 1)
    (3.087, 1) (3.429, 1)
};

% ===== GEMINI TEXT BOXPLOT =====
\addplot+[
    boxplot prepared={
        lower whisker=1.0000,
        lower quartile=1.0005,
        median=1.0020,
        upper quartile=1.0083,
        upper whisker=1.0199,
    },
    boxplot/draw position=2,
    boxplot/box extend=0.6,
    fill=textorange!25,
    draw=textorange,
    thick,
] coordinates {};

% Gemini Text outliers (164 values)
\addplot[only marks, mark=*, mark size=1.8pt, textorange, opacity=0.2] coordinates {
    (1.020, 2) (1.020, 2) (1.021, 2) (1.021, 2) (1.021, 2) (1.021, 2) (1.021, 2) (1.021, 2) (1.022, 2) (1.022, 2)
    (1.022, 2) (1.023, 2) (1.023, 2) (1.023, 2) (1.024, 2) (1.024, 2) (1.024, 2) (1.025, 2) (1.025, 2) (1.025, 2)
    (1.025, 2) (1.025, 2) (1.026, 2) (1.026, 2) (1.026, 2) (1.026, 2) (1.027, 2) (1.027, 2) (1.028, 2) (1.028, 2)
    (1.028, 2) (1.029, 2) (1.029, 2) (1.030, 2) (1.030, 2) (1.031, 2) (1.031, 2) (1.032, 2) (1.032, 2) (1.032, 2)
    (1.033, 2) (1.035, 2) (1.035, 2) (1.036, 2) (1.037, 2) (1.037, 2) (1.037, 2) (1.037, 2) (1.038, 2) (1.038, 2)
    (1.038, 2) (1.039, 2) (1.039, 2) (1.039, 2) (1.040, 2) (1.040, 2) (1.040, 2) (1.041, 2) (1.041, 2) (1.041, 2)
    (1.042, 2) (1.042, 2) (1.043, 2) (1.044, 2) (1.044, 2) (1.044, 2) (1.045, 2) (1.045, 2) (1.045, 2) (1.047, 2)
    (1.047, 2) (1.047, 2) (1.048, 2) (1.048, 2) (1.050, 2) (1.053, 2) (1.053, 2) (1.054, 2) (1.054, 2) (1.055, 2)
    (1.056, 2) (1.057, 2) (1.058, 2) (1.059, 2) (1.061, 2) (1.061, 2) (1.062, 2) (1.062, 2) (1.063, 2) (1.065, 2)
    (1.067, 2) (1.067, 2) (1.071, 2) (1.074, 2) (1.076, 2) (1.077, 2) (1.080, 2) (1.080, 2) (1.080, 2) (1.081, 2)
    (1.082, 2) (1.083, 2) (1.086, 2) (1.087, 2) (1.088, 2) (1.089, 2) (1.090, 2) (1.099, 2) (1.120, 2) (1.122, 2)
    (1.124, 2) (1.125, 2) (1.126, 2) (1.127, 2) (1.129, 2) (1.132, 2) (1.141, 2) (1.141, 2) (1.144, 2) (1.145, 2)
    (1.158, 2) (1.166, 2) (1.169, 2) (1.172, 2) (1.181, 2) (1.181, 2) (1.182, 2) (1.182, 2) (1.187, 2) (1.202, 2)
    (1.212, 2) (1.233, 2) (1.234, 2) (1.235, 2) (1.242, 2) (1.254, 2) (1.254, 2) (1.258, 2) (1.267, 2) (1.270, 2)
    (1.284, 2) (1.293, 2) (1.299, 2) (1.327, 2) (1.332, 2) (1.341, 2) (1.364, 2) (1.364, 2) (1.401, 2) (1.424, 2)
    (1.455, 2) (1.487, 2) (1.500, 2) (1.548, 2) (1.601, 2) (1.703, 2) (1.743, 2) (1.910, 2) (2.008, 2) (2.016, 2)
    (2.023, 2) (2.143, 2) (2.343, 2) (4.000, 2)
};

% Reference line
\draw[gray!60, densely dashed, thin] (1.0, 0.4) -- (1.0, 2.6);

\end{axis}

% ===== QWEN PLOT (BOTTOM) =====
\begin{axis}[
    name=qwen,
    at={(gemini.south west)},
    anchor=north west,
    yshift=-1.2cm,
    width=10cm,
    height=3cm,
    title={\textbf{Qwen2.5-72B}},
    title style={font=\small, yshift=-2mm},
    xlabel={Perplexity},
    xlabel style={font=\small},
    ytick={1, 2},
    yticklabels={Visual, Text},
    yticklabel style={font=\small},
    tick label style={font=\footnotesize},
    xmin=0.9,
    xmax=4.2,
    ymin=0.5,
    ymax=2.5,
    xmajorgrids=true,
    ymajorgrids=false,
    grid style={line width=0.2pt, draw=gray!15},
    axis lines=left,
    axis line style={gray!70},
    tick style={gray!70},
    enlarge y limits=0.15,
    boxplot/draw direction=x,
]

% ===== QWEN VISUAL BOXPLOT =====
\addplot+[
    boxplot prepared={
        lower whisker=2.2232,
        lower quartile=3.2110,
        median=3.6418,
        upper quartile=3.8696,
        upper whisker=4.0000,
    },
    boxplot/draw position=1,
    boxplot/box extend=0.6,
    fill=visualblue!25,
    draw=visualblue,
    thick,
] coordinates {};

% Qwen Visual outliers (43 values)
\addplot[only marks, mark=*, mark size=1.8pt, visualblue, opacity=0.4] coordinates {
    (1.135, 1) (1.155, 1) (1.172, 1) (1.195, 1) (1.214, 1) (1.251, 1) (1.379, 1) (1.448, 1) (1.481, 1) (1.502, 1)
    (1.532, 1) (1.583, 1) (1.613, 1) (1.617, 1) (1.662, 1) (1.676, 1) (1.692, 1) (1.742, 1) (1.816, 1) (1.832, 1)
    (1.835, 1) (1.855, 1) (1.855, 1) (1.869, 1) (1.896, 1) (1.903, 1) (1.911, 1) (1.932, 1) (1.975, 1) (1.984, 1)
    (1.992, 1) (2.015, 1) (2.048, 1) (2.059, 1) (2.082, 1) (2.113, 1) (2.150, 1) (2.155, 1) (2.166, 1) (2.168, 1)
    (2.198, 1) (2.204, 1) (2.215, 1)
};

% ===== QWEN TEXT BOXPLOT =====
\addplot+[
    boxplot prepared={
        lower whisker=1.0018,
        lower quartile=1.3979,
        median=2.0436,
        upper quartile=2.7947,
        upper whisker=3.9612,
    },
    boxplot/draw position=2,
    boxplot/box extend=0.6,
    fill=textorange!25,
    draw=textorange,
    thick,
] coordinates {};

% Qwen Text outliers (0 values - none)

% Reference line
\draw[gray!60, densely dashed, thin] (1.0, 0.4) -- (1.0, 2.6);

\end{axis}

% % ===== LEGEND =====
% \node[draw=gray!50, rounded corners=3pt, inner sep=6pt, anchor=north] at (4.5, -4.8) {
%     \footnotesize
%     \tikz\fill[visualblue!25, draw=visualblue, thick] (0,0) rectangle (0.3,0.25); Visual
%     \hspace{1em}
%     \tikz\fill[textorange!25, draw=textorange, thick] (0,0) rectangle (0.3,0.25); Text
% };

\end{tikzpicture}}
    \caption{\small Uncertainty analysis comparing Gemini2.5-Pro and Qwen2.5-72B on text-option vs. visual-option questions. Perplexity score 1=absolute confidence, 4=absolute uncertainty. The boxplots of perplexity score reveal a stark contrast: Gemini is overconfident in its predictions regardless of the evaluation modality, while Qwen is more uncertain on both tasks, more uncertain on visual information than text. Gemini makes no such distinctions and confidently predicts the wrong answer.}
    \label{fig:uncertainty_analysis}
    \vspace{-3mm}
\end{figure}
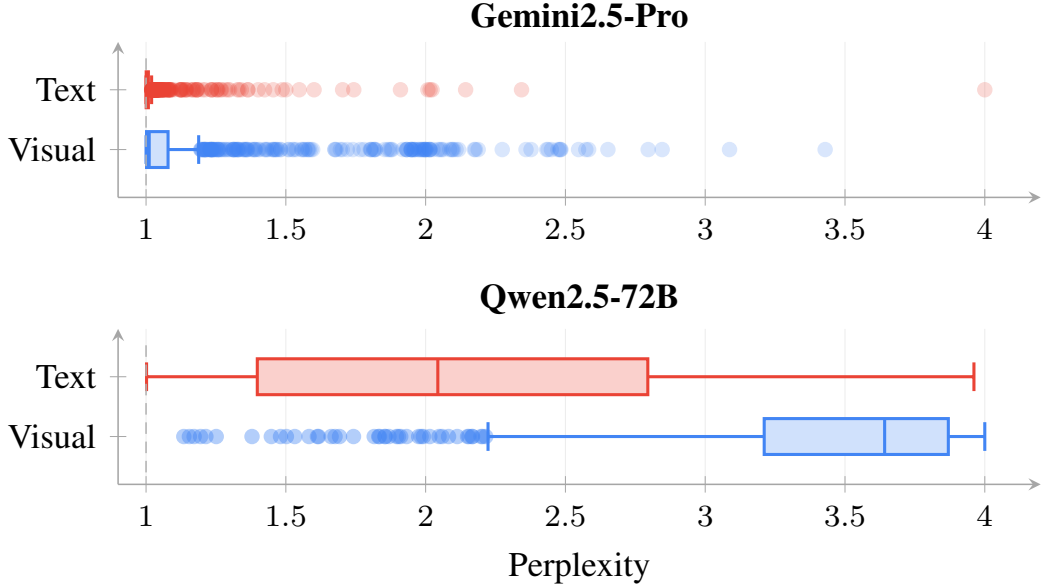

For each question, we treat the four answer options as a categorical distribution induced by the model's logits and estimate model uncertainty over the options via perplexity~\cite{xia2025survey, xiong2024efficient}. By comparing these quantities across text-option vs. visual-option settings on the Gemini 2.5 Pro  and Qwen2.5-72B responses in~\Cref{fig:uncertainty_analysis}, we find a stark contrast:
% we aim to understand whether the modality shift changes the internal behavior of the model, i.e., whether failure on video options comes with appropriate caution or systematically overconfident behavior.
we find that Gemini is overall more overconfident in its predictions, and does not calibrate the difference in evaluation modality (text and vision), i.e., there is a minimal gap in its uncertainty among the two modalities. Its predictions are overconfident with respect to both. Qwen, on the other hand, is more uncertain on both modalities (is not overconfident in either), and calibrated among modalities as well; it is much more uncertain on visual data than text. These are interesting results, as both models have low accuracies on the temporal frame retrieval task (Table \ref{tab:main_results}), which allows us to frame their behavior in terms of confidence and calibration. 

\section{Dataset expansion roadmap}

In the CMD annotation pipeline, the real time constraint is the selection of suitable video candidates - finding a usable video requires a human to judge whether the contact is unambiguous, whether the post-contact trajectory is visible and long enough to imply the outcome, and whether the scene contains causally unrelated footage. These issues were not encountered by the original study \cite{Strickland2011EventCE}, as there were only six videos in their work.

But once these candidates are identified, the annotation cost itself from the human is not expensive - per source video the annotator marks four things: the sub-range containing one target action, the contact/release frame (the probe frame at test), the interval to exercise in the incomplete conditions, and an interval of unrelated same-scene footage. Through these annotations, the tool can synthesize the different stimuli. Marking a selected video takes 5–10 minutes.

Therefore, in the dataset expansion process, the crowd workers would first need to filter out candidates based on the judgment described above, and only then engage in the annotation. Once the first step is achieved, the second step is quicker. We plan to engage crowdworkers in this process. We have also released the annotation tool as part of our code release.

The TFR and VFP tasks are relatively straightforward to expand. Both scripts automate the frame selection and generation process, while for VFP, potentially infinite videos could be generated as it is a 3D object simulator.

For all tasks, there is dataset flexibility - Something-Something V2, EPIC-KITCHENS, Ego4D, are some examples of other source datasets from which we can extract frames.

% \section{Prompt sensitivity}
% We run an ablation for Gemini and Kimi K2.6 on TFR using prompt variations. We modify the original prompt's wording and structure, but keep the semantics unchanged. If prompt sensitivity were the key differentiator, we would observe a significant difference in the vision-text performance gap. However, consistent with our central claim, we observe a minimal difference in the vision-text performance gap, demonstrated in \cref{tab:text-visual-performance}. It cannot, therefore, be attributed to variations in the prompt.

% \begin{table}[h]
% \centering
% \caption{\small Text and visual performance across model variants. $\Delta$ denotes the difference between visual and text performance.}
% \footnotesize
% \begin{tabular}{lccc}
% \toprule
% Variant & Text & Visual & $\Delta$ \\
% \midrule
% Gemini Original & 0.739 & 0.307 & $-0.432$ \\
% Gemini Modified & 0.744 & 0.306 & $-0.438$ \\
% Kimi Original   & 0.721 & 0.334 & $-0.386$ \\
% Kimi Modified   & 0.714 & 0.256 & $-0.458$ \\
% \bottomrule
% \end{tabular}
% \label{tab:text-visual-performance}
% \end{table}

% This is consistent with our central claim in the paper - visual querying surfaces gaps that text querying alone cannot elucidate in video understanding. 

% \newpage
% \input{checklist.tex}

%%%%%%%%%%%%%%%%%%%%%%%%%%%%%%%%%%%%%%%%%%%%%%%%%%%%%%%%%%%%

\end{document}